\RequirePackage{fix-cm}
\documentclass[twocolumn]{svjour3}          
\smartqed  
\usepackage{graphicx}
\usepackage{xcolor}            
\usepackage{amsmath,amssymb,amsfonts}
\usepackage{mathtools}         
\usepackage{booktabs}
\usepackage{multirow}
\usepackage{arydshln}
\usepackage{textcomp}
\usepackage{textgreek}
\usepackage{colortbl}

\usepackage{algorithm}
\usepackage{algorithmicx}
\usepackage{algpseudocode}

\usepackage{minted}            

\usepackage{caption}
\usepackage{subcaption}

\usepackage[square,numbers]{natbib}   
\usepackage{hyperref}          

\usepackage{silence}
\usepackage{lineno}
\usepackage{manyfoot}
\usepackage{mathrsfs}
\usepackage[title]{appendix}
\usepackage{color}             
\usepackage{comment}
\usepackage{academicons}
\usepackage{orcidlink}

\begin{document}

\title{Neural Plasticity-Inspired Multimodal Foundation Model for Earth Observation 
}


\author{Zhitong Xiong \and
        Yi Wang \and
        Fahong Zhang \and
        Adam J. Stewart \and
        Joëlle Hanna \and
        Damian Borth \and
        Ioannis Papoutsis \and
        Bertrand Le Saux \and
        Gustau Camps-Valls \and
        Xiao Xiang Zhu*}


\institute{
Z. Xiong \and Y. Wang \and F. Zhang \and A. J. Stewart \and X. X. Zhu \at
Chair of Data Science in Earth Observation, Technical University of Munich (TUM), Arcisstraße 21, 80333 Munich, Germany
\and
X. X. Zhu (\textbf{Corresponding Author}) \at
Munich Center for Machine Learning, 80333 Munich, Germany \\
\email{xiaoxiang.zhu@tum.de}
\and
J. Hanna \and D. Borth \at
AIML Lab, School of Computer Science, University of St. Gallen, Rosenbergstrasse 30, 9000 St. Gallen, Switzerland
\and
I. Papoutsis \at
School of Rural, Surveying and Geoinformatics Engineering, National Technical University of Athens, 157 73 Zografou, Greece
\and
B. Le Saux \at
ESRIN, \textPhi-lab, European Space Agency (ESA), 00044 Frascati, Italy
\and
G. Camps-Valls \at
Image Processing Laboratory (IPL), Universitat de València, Cat. Agustín Escardino Benlloch 9, 46980 Paterna, València, Spain
}

\date{Received: date / Accepted: date}

\maketitle

\begin{abstract}
Earth observation (EO) in open-world settings presents a unique challenge: different applications rely on diverse sensor modalities, each with varying ground sampling distances, spectral ranges, and numbers of spectral bands. However, existing EO foundation models are typically tailored to specific sensor types, making them inflexible when generalizing across the heterogeneous landscape of EO data. To address this, we propose the Dynamic One-For-All (DOFA) model, a unified, multimodal foundation framework designed for diverse vision tasks in EO. Inspired by neural plasticity, DOFA utilizes a wavelength-conditioned dynamic hypernetwork to process inputs from five distinct satellite sensors flexibly. By continually pretraining on five EO modalities, DOFA achieves state-of-the-art performance across multiple downstream tasks and generalizes well to unseen modalities. Enhanced with hybrid continual pretraining, DOFA+ requires significantly fewer computational resources while outperforming counterparts trained with extensive GPU budgets. Experiments on diverse datasets highlight DOFA's potential as a foundation for general-purpose vision models in the sensor-diverse EO domain. The code and pre-trained weights are publicly available at \url{https://github.com/zhu-xlab/DOFA}.

\end{abstract}

\section{Introduction}
\label{intro}

\begin{figure}
    \centering
    \includegraphics[width=0.48\textwidth]{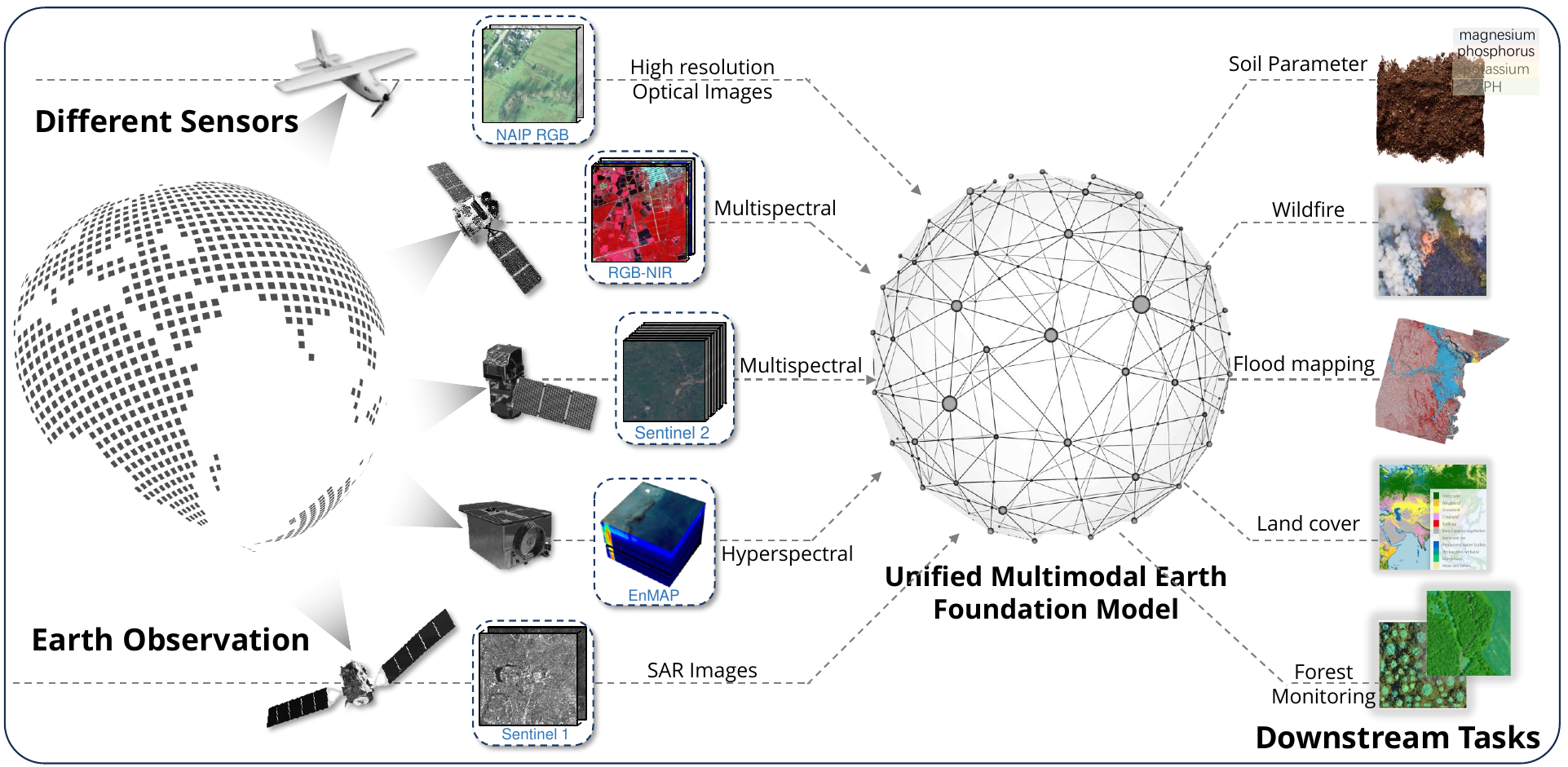}
    \caption{\textbf{Motivation of DOFA.} Our primary purpose is to develop versatile foundation models capable of adaptively processing various EO data modalities.}
    \label{fig::motiv}
\end{figure}

Earth observation (EO) through satellite remote sensing provides unparalleled opportunities for modeling Earth's surface dynamics, addressing global challenges such as climate change, urbanization, and biodiversity loss~\cite{camps2011remote,camps2021deep,reichstein2019deep}. Core vision tasks like classification, detection, and segmentation support critical applications from disaster response to resource management. Achieving this requires integrating data from diverse sensors with varying spectral and spatial characteristics, making EO an open-world visual understanding problem. Foundation models (FMs) offer a scalable solution by enabling generalization across modalities, reducing task-specific supervision, and serving as a unified backbone for downstream EO tasks, as illustrated in Fig.~\ref{fig::motiv}.

Existing FMs, though promising, remain rigid and inflexible, typically pretrained using fixed spectral band configurations or specialized for individual EO modalities. For instance, models such as GFM~\cite{mendieta2023towards}, Scale-MAE~\cite{reed2023scale}, SatMAE~\cite{cong2022satmae}, CROMA~\cite{fuller2023croma}, and SpectralGPT~\cite{hong2024spectralgpt} illustrate these limitations. This severely restricts their adaptability in dynamic real-world EO scenarios, where new sensors and spectral band configurations continuously emerge. Consequently, extensive computational and human resources are required to adapt them to unseen sensors and spectral combinations. In summary, existing approaches that develop separate foundation models or use isolated visual encoders for multi-sensor data fail to capture inter-sensor relationships, resulting in key limitations:
\begin{enumerate}
    \item The learned representation may not effectively capture such an intersensor relationship.
    \item The performance of FMs will degrade when downstream tasks utilize data from unseen sensors with varying numbers of spectral bands.
    \item The development of individual FMs requires considerably more computing resources and is not flexible in real-world applications.
\end{enumerate}
Addressing this critical limitation aligns with a central theme of open-world EO visual understanding: developing FMs capable of dynamically and efficiently adapting to downstream applications with flexible input data modalities.

\begin{figure}
    \centering
    \includegraphics[width=0.48\textwidth]{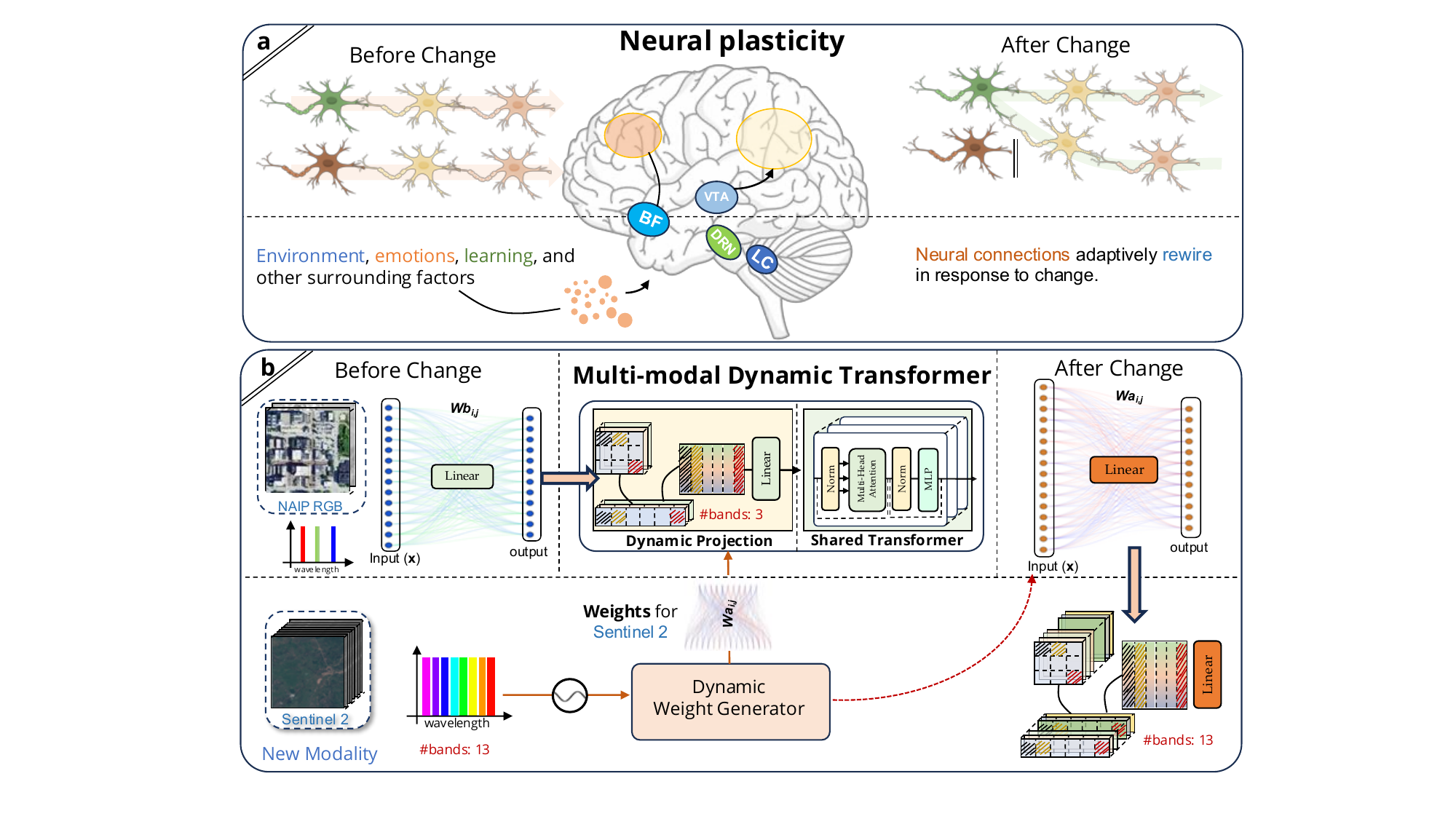}
    \caption{\textbf{Motivation and main architecture of DOFA.} We design DOFA to emulate the Neuroplasticity~\cite{hebb2005organization,zucker2002short,dan2004spike} mechanism for processing multimodal EO data. \textbf{(1)} Illustration of the brain's capability to adapt its structure and function to learned information, experience, or injury. \textbf{(2)} Illustration of the core idea: DOFA is designed to adaptively alter its network weights in response to novel data modalities.}
    \label{fig::overall}
\end{figure}

In response to this challenge, we propose the Dynamic One-For-All (DOFA) model, a versatile, adaptive multimodal foundation model explicitly designed for the EO domain. Inspired by neuroplasticity, the brain's dynamic reorganization capacity in response to novel stimuli~\cite{dayan2011neuroplasticity,lillicrap2020backpropagation}, DOFA integrates a wavelength-conditioned dynamic hypernetwork within a unified Transformer architecture, as shown in Fig.~\ref{fig::overall}. This enables DOFA to flexibly accommodate varying spectral bands and sensor modalities, including those unseen during initial pre-training. Specifically, DOFA utilizes wavelength as a unifying parameter across various EO modalities to achieve a more cohesive multimodal representation. At its core, the model integrates a hypernetwork~\cite{hypernetworks} that dynamically generates network weights based on the central wavelengths of each spectral band. This dynamic weight generator adjusts network weights to align with the specific modality of the input data, facilitating a customized network for each modality. Additionally, DOFA integrates a shared vision backbone, acting as a universal feature learning module for all heterogeneous data modalities. 

Pretraining unified FMs across diverse EO modalities presents a significant challenge due to the need for substantial computational resources. To address this, DOFA adopts continuous pretraining via masked image modeling (MIM) and knowledge distillation, significantly reducing computational cost. Building on this, we further propose DOFA+, which is initialized from a powerful open-source pre-trained vision model (DINOv2~\cite{oquab2023dinov2}) and applies the MIM objective to enable efficient domain adaptation, without altering the underlying architecture. Through a hierarchical distillation strategy, it preserves strong semantic priors from the source model while guiding the learning of EO-specific visual patterns through local reconstruction. With parameter-efficient fine-tuning, DOFA and DOFA+ enable rapid, label-efficient adaptation to a wide range of multimodal EO tasks, including image classification, semantic segmentation, object detection, and environmental change detection.

Our contributions can be summarized as follows:
\begin{enumerate}
    \item We introduce DOFA, a neuroplasticity-inspired architecture that uses wavelength as a unifying input across EO modalities. A wavelength-conditioned dynamic hypernetwork enables flexible adaptation to varying and unseen spectral bands within a single unified Transformer framework.
    \item The proposed hypernetwork-based architecture enables efficient continual pretraining across diverse EO modalities by interpolating in weight space according to wavelength configurations. Together with wavelength-aware MIM and feature distillation, this enables efficient EO domain adaptation with minimal data and compute.
    \item We further introduce DOFA+. DOFA+ seeds itself with a strong vision prior and then employs a dual training strategy: (i) wavelength-aware MIM to capture EO-specific spatial patterns, and (ii) hierarchical feature distillation to align and refine the inherited semantic representations from the vision prior.
    \item Extensive experiments demonstrate that DOFA and DOFA+ achieve state-of-the-art performance across a wide range of EO tasks and modalities. Our models generalize well to unseen sensors and diverse spectral configurations without retraining, offering greater flexibility in open-world settings with reduced computational costs.
\end{enumerate}
Extensive experiments across 20+ EO tasks validate DOFA's robust adaptability, efficiency, and scalability. Our approach represents a meaningful advancement toward open-world EO visual understanding, emphasizing continual multimodal learning and sustainable adaptability within dynamic and heterogeneous real-world EO environments.

\section{Related Work}
\label{sec:rw}
Early efforts to develop EO foundation models were devoted to generating effective embeddings for data from a single modality. For example, SeCo~\cite{manas2021seasonal} and CACo~\cite{mall2023change} leverage temporal information from acquired images to learn temporal-sensitive and temporal-invariant feature representations. GFM~\cite{mendieta2023towards} devises a continual pretraining paradigm that leverages ImageNet pretrained features to accelerate model convergence on EO data. Cha et al.~\cite{cha2023billion} explore the impact of scaling up the number of parameters in foundation models, specifically on Google Earth images. 
Another line of research addresses the adaptability of feature representations across EO data with different ground sample distances (GSD).

RingMo~\cite{yao2023ringmo} introduces a patch-incomplete mask strategy during the masked image modeling phase, preventing the oversight of small objects within a single patch.
Scale-MAE~\cite{reed2023scale} takes a different approach by substituting the positional encoding within ViT~\cite{dosovitskiy2020image} with a GSD positional encoding, incorporating GSD information into the representation learning process. USat~\cite{irvin2023usat} adopts a strategy of encoding a higher number of patches for bands with lower GSD and a lower number of patches for bands with higher GSD.
Another significant research question is how to achieve a unified representation for different modalities, such as RGB, multispectral, hyperspectral, and radar data.
In this regard, SSL4EO-S12~\cite{wang2023ssl4eo} integrates the features from multispectral and SAR modalities using an early fusion strategy. SatMAE~\cite{cong2022satmae} suggests grouping subsets of spectral bands and adding a spectral encoding to each spectral group. 

Real-world data are characterized by diverse modalities, including but not limited to images, videos, text, audio, depth information, and point clouds. 
The capacity of foundation models to effectively handle this variety of downstream tasks hinges on their ability to process multimodal data. In this context, OFA-Net~\cite{OFA} proposes using modality-specific patch embedding layers to learn unified representations across diverse EO modalities. It further demonstrates that a single shared Transformer backbone is both sufficient and effective for capturing generalizable representations spanning multiple types of EO data.
CROMA~\cite{fuller2023croma} first develops two unimodal encoders to encode multispectral and SAR data individually. Subsequently, it utilizes a cross-modal radar-optical Transformer that leverages cross-attention to extract the unified representation.
DeCUR~\cite{wang2023decur} is a bi-modal self-supervised foundation model that decouples the unique and common representations between the two modalities. 

SpectralGPT~\cite{hong2024spectralgpt} is a foundation model meticulously tailored for hyperspectral remote sensing data. It designs a 3D masking strategy, an encoder for learning representations from spatial-spectral mixed tokens, and a decoder with multi-target reconstruction to preserve spectral characteristics. Beyond these, efforts have also been directed towards encoding geo-locational information into the feature representation.
Notable examples include GASSL~\cite{ayush2021geography}, GeoCLIP~\cite{cepeda2023geoclip}, SatCLIP~\cite{klemmer2023satclip}, SkySense~\cite{guo2023skysense} and Tile2Vec~\cite{jean2019tile2vec}. 

Recent efforts aim to handle the full diversity of EO data in a single backbone. SkySense V2~\cite{zhang2025skysensev2} unifies optical, SAR, and elevation inputs with modality-prompt tokens and adaptive patch merging, while AnySat~\cite{astruc2025anysat} is a multimodal model based on joint embedding predictive architecture and scale-adaptive spatial encoders. Panopticon~\cite{waldmann2025panopticon} treats co-located multi-sensor images as natural augmentations, adding cross-channel attention to remain sensor-agnostic.  Galileo~\cite{tseng2025galileo} couples global masked modeling with local contrastive objectives, yielding a generalist model that outperforms task-specific baselines. However, for existing models, it is tricky to handle situations where the number of spectral bands changes in downstream tasks without retraining the models. DOFA and DOFA+ overcome this by employing a dynamic weight generator to encode spectral bands into dynamic weights for deep representation learning.

\begin{figure*}[htbp]
    \centering
    \includegraphics[width=0.95\textwidth]{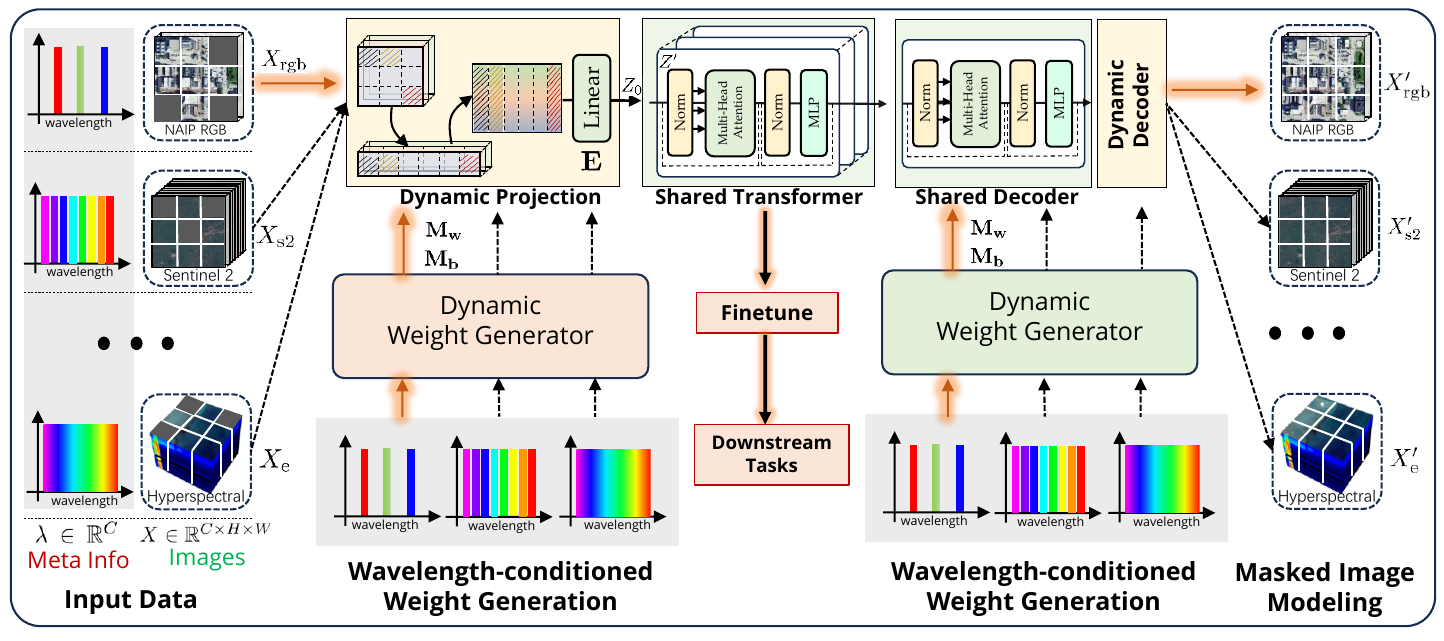}
    \caption{\textbf{Architecture and training details.} DOFA builds on masked image modeling, introducing a significant advancement by processing input images with any number of channels within a single framework.}
    \label{ooarc}
\end{figure*}

\section{Methodology}
\label{sec:2}
Here, we provide detailed information about the proposed DOFA and DOFA+ models, along with a more detailed presentation of the training method.

\subsection{Preliminary}
\label{preliminary}
Given an input image \( \mathbf{X} \in \mathbb{R}^{C\times H \times W} \), where \( H \), \( W \), and \( C \) represent the height, width, and number of channels, respectively, the image is first divided into a patch sequence. Each patch has a fixed spatial size \( P \times P \) with $C$ channels, and thus the image is converted into \( N = \frac{HW}{P^2} \) patches. Each patch is flattened into a vector and linearly transformed into a \( D \)-dimensional embedding. This transformation is represented by a trainable embedding matrix \( \mathbf{E} \in \mathbb{R}^{P^2 C \times D} \). 

Formally, the patch embedding can be described as
\begin{equation}
\mathbf{X} = [\mathbf{X}_{p_1}; \mathbf{X}_{p_2}; \ldots ; \mathbf{X}_{p_N}], \quad \mathbf{X}_{p_i} \in \mathbb{R}^{P^2 C},
\end{equation}
where \( \mathbf{X}_{p_i} \) is the flattened vector of the \( i \)-th patch. Next, the flattened vectors are linearly projected into $D-$dimensional embeddings with a learnable embedding matrix:
\begin{equation}
\mathbf{Z_0} = [\mathbf{X}_{p_1}\mathbf{E}; \mathbf{X}_{p_2}\mathbf{E}; \ldots ; \mathbf{X}_{p_N}\mathbf{E}], \quad \mathbf{Z_0} \in \mathbb{R}^{N \times D},
\end{equation}
where \( \mathbf{Z}_0 \) represents the sequence of patch embeddings. Note that this process can be implemented utilizing a single convolution layer with a \(P\times P\) kernel, \(C\) input channels, and \(D\) output channels.
Class token \( \mathbf{X}_\text{cls} \), an additional learnable embedding, is prepended to the sequence. Finally, position embeddings are added to retain positional information.
\begin{equation}
\mathbf{Z'} = [\mathbf{X}_\text{cls}; \mathbf{Z_0}] + \mathbf{E_\text{pos}}, \quad \mathbf{Z'} \in \mathbb{R}^{(N+1) \times D}.
\end{equation}
Here, \( \mathbf{E_{pos}} \) denotes the position embeddings, and the resulting \( \mathbf{Z'} \) serves as the input to the subsequent layers of the ViT architecture. 

\subsection{Architecture overview}
The patch embedding layer transforms the input image into a sequence of embeddings that the self-attention mechanism of the Transformer can process. A straightforward way to handle the input data from different modalities is to utilize multiple patch embedding layers to convert data with different spectral wavelengths into embeddings with the same dimension~\cite{OFA}. Suppose that the input image \( \mathbf{X} \) of dimensions \( \mathbb{R}^{C \times H \times W} \) can originate from various data modalities. Initially, images from different sources are standardized to height $H$ and width $W$. Specifically, we consider five distinct modalities: Sentinel-1 data (\( \mathbf{X}_\text{s1} \)) with two SAR channels (\( \mathbb{R}^{2 \times H \times W} \)), Sentinel 2 data (\( \mathbf{X}_\text{s2} \)) with nine multispectral channels (\( \mathbb{R}^{9 \times H \times W} \)),  Gaofen data (\( \mathbf{X}_\text{g} \)) with four multispectral channels (\( \mathbb{R}^{4 \times H \times W} \)), NAIP imagery (\( \mathbf{X}_\text{rgb} \)) with three RGB channels (\( \mathbb{R}^{3 \times H \times W} \)), and EnMAP data (\( \mathbf{X}_\text{e} \)) with 202 available hyperspectral channels (\( \mathbb{R}^{202 \times H \times W} \)). Note that, for the sake of simplicity, we omit the batch size from the notation of tensors. 
\begin{figure}
    \centering
    \includegraphics[width=0.48\textwidth]{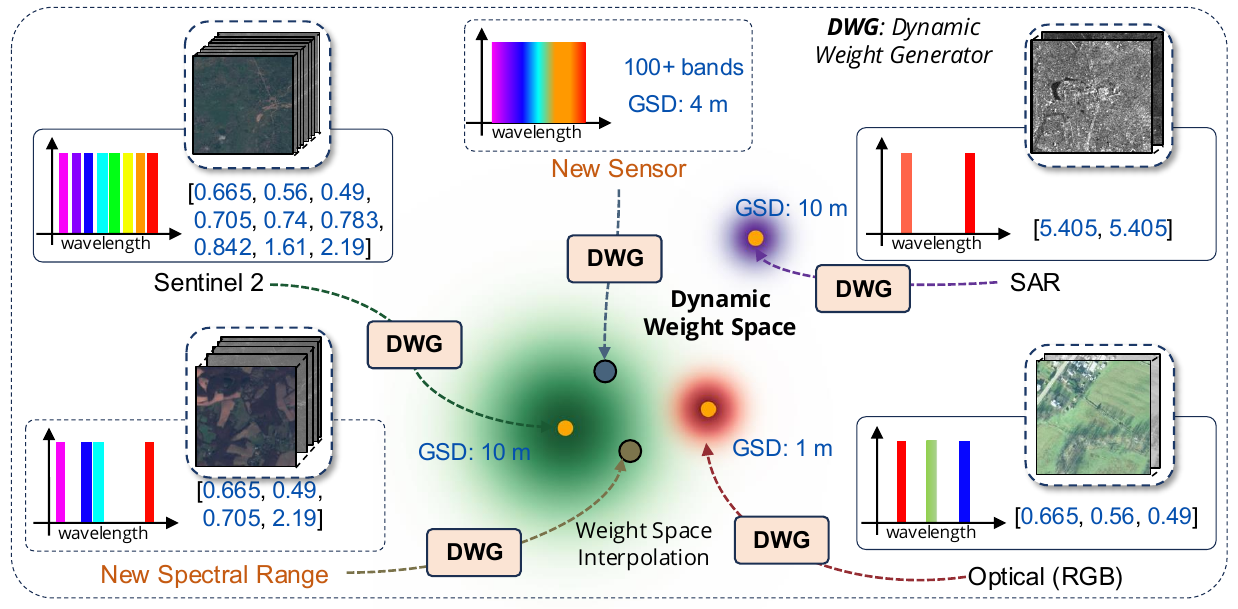}
    \caption{Illustration of weight space interpolation for new sensors or a combination of spectral bands.}
    \label{wave_vis}
\end{figure}

As illustrated in Fig.~\ref{ooarc}, the whole architecture follows the design of masked image modeling (MIM)~\cite{he2022masked}. The main difference from traditional masked autoencoders (MAE) lies in DOFA's capacity to process input images with various channels. This flexibility is achieved through a hypernetwork-based dynamic weight generator. The dynamic weight generator takes inputs from the spectral wavelength associated with each image channel, and dynamically predicts the patch embedding matrix $\mathbf{E}$ for different data modalities. As presented in Fig.~\ref{wave_vis}, the dynamic weight generator maps spectral band configurations to a weight space. In open-world tasks with new spectral ranges or sensors, it interpolates this space to produce appropriate weights for the given input. The latent representations are then passed through a series of shared Transformer blocks for learning generalizable multimodal representations.
\begin{figure*}[htbp]
    \centering
    \includegraphics[width=0.8\textwidth]{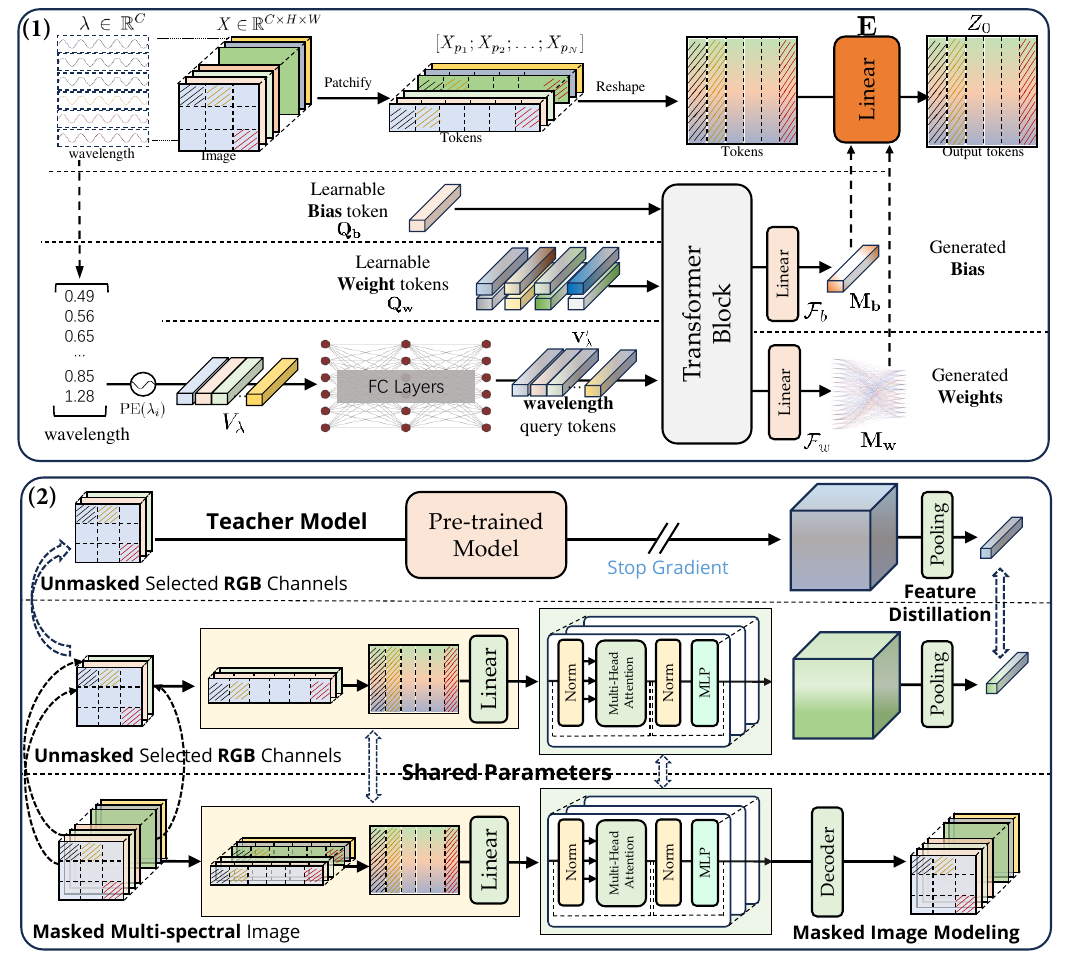}
    \caption{\textbf{Dynamic weight generator and continual training framework.} (1) The central wavelengths of each band are utilized to derive weights tailored to each wavelength. (2) Continual pretraining process. There is a distillation and a reconstruction loss.}
    \label{DPE}
\end{figure*}

Parallel to the dynamic weight generation for the encoder part of the network, the dynamic decoder is responsible for reconstructing the output image from the encoded latent space. Similarly, the dynamic decoder utilizes another set of dynamically generated weights to ensure that the reconstructed image matches the number of spectral bands of the target modality. We employ a MIM strategy to train this self-supervised architecture. The input images are masked randomly, and the model learns to reconstruct these missing parts. As the parameters in DOFA are learned across different modalities, this process helps the model to learn robust multimodal representations beneficial for various EO tasks. After the pre-training process, the model can be transferred to various EO applications without extensive retraining thanks to the dynamic weight generation, from high-resolution optical imaging to multispectral and hyperspectral sensing.

\subsection{Wavelength-conditioned dynamic patch embedding}
\label{WCDPE}
To manage the diversity of spectral bands across different modalities, we project the data into a latent space with uniform feature dimensionality using the dynamic patch embedding layer $\mathcal{F}_\text{dpe}$. As described before, we denote the input image as $\mathbf{X}\in \mathbb{R}^{C\times H\times W}$. Fig.~\ref{DPE} (1) illustrates the detailed steps used to compute the dynamic weights given the wavelength information of each channel. Each channel of the input image has a corresponding central wavelength. The wavelengths of an input image with $C$ channels can be represented by $\mathbf{\lambda} \in \mathbb{R}^{C}$. To convert the wavelengths to a higher-dimensional feature space, we encode the wavelengths $\lambda$ using a 1D sine-cosine positional encoding:
\begin{equation}
\mathbf{V_{\lambda}} = \text{PE}(\mathbf{\lambda}) \in \mathbb{R}^{C \times D_{\lambda}},    
\end{equation}
where $D_{\lambda}$ is the dimension of the converted wavelength feature.
The positional encoding $\text{PE}(\lambda_i)$ for wavelength $\lambda_i$ in channel $i$ is given by:
\begin{equation}
\begin{aligned}
\text{PE}(\lambda_i, 2k) &= \sin\left(\frac{\lambda_i}{10000^{2k / D_{\lambda}}}\right), \\ 
\text{PE}(\lambda_i, 2k+1) &= \cos\left(\frac{\lambda_i}{10000^{2k / D_{\lambda}}}\right),
\end{aligned}
\end{equation}
where $k = 0, \ldots, \frac{D_{\lambda}}{2} - 1$. The positionally encoded wavelengths $\mathbf{V}_{\lambda}$ are further transformed through two fully-connected layers with residual connections:
\begin{equation}
\mathbf{\mathbf{V'}_{\lambda}} = \text{ReLU}(\mathcal{F}_2(\text{ReLU}(\mathcal{F}_1(\mathbf{V}_{\lambda})))) + \mathbf{V}_{\lambda},
\end{equation}
where $\mathcal{F}_1$ and $\mathcal{F}_2$ represent the fully-connected layers, and ReLU denotes the Rectified Linear Unit activation function~\cite{agarap2018deep}.

Next, we employ a Transformer encoder~\cite{vaswani2017attention} layer with four attention heads to generate the dynamic weights and bias for each wavelength. Specifically, the embedding $\mathbf{V'_{\lambda}}$, $N_w$ learnable query tokens $\mathbf{Q_w}$, and one learnable bias query token $\mathbf{Q_b}$ are concatenated together to form the input to the Transformer encoder:
\begin{equation}
\mathbf{V''} = \text{TransformerEncoder}(\text{Concat}(\mathbf{Q_w}, \mathbf{V'}_{\lambda}, \mathbf{Q_b})).
\end{equation}
We subsequently extract the embeddings $\mathbf{V''}_{w}$ that correspond to the weight query tokens $\mathbf{Q}_w$ from $\mathbf{V''}$, as well as the embeddings $\mathbf{V''}_{b}$ associated with the bias query tokens $\mathbf{Q_b}$ from $\mathbf{V''}$. Then, two fully-connected layers are utilized to generate the dynamic weights and biases:
\begin{equation}
\begin{aligned}
\mathbf{M_w} &= \mathcal{F}_w(\mathbf{V''}_{w} + \mathbf{V'}_{\lambda}) \in \mathbb{R}^{C \times P^2 D}, \\
\mathbf{M_b} &= \mathcal{F}_b(\mathbf{V''}_{b}) \in \mathbb{R}^{C \times D},
\end{aligned}
\end{equation}
where $\mathcal{F}_w$ and $\mathcal{F}_b$ denote the fully-connected layers for weight and bias generation, respectively. As introduced in the Mathematical formalism section, the patch embedding layer can be implemented efficiently using a convolution layer. Thus, we reshape the generated weights into the convolution kernel as:
\begin{equation}
\mathbf{K_{\text{conv}}} = \text{Reshape}(\mathbf{M_w}, [D, C, P, P]),
\end{equation}
The convolution operation for patch embedding is then performed using the dynamically generated weights $\mathbf{K_{\text{conv}}}$ and biases $\mathbf{M_b}$:
\begin{equation}
\text{PatchEmbedding} := \text{Conv}(\mathbf{X}, \mathbf{K_{\text{conv}}}, \mathbf{M_b}).
\end{equation}
where $\text{Conv}$ denotes the convolution operation.
Utilizing this approach, the patch embedding layer achieves independence from the number of spectral bands of the input images. The weights for these layers are dynamically generated based on the central wavelength of each channel in a compositional manner. This mechanism enables the model to learn modality-specific representations dynamically, thereby enhancing its adaptability and performance across various data domains.

For the wavelength-conditioned dynamic decoder, we use different parameters to generate dynamic weights and biases. The computation process is similar to the dynamic patch embedding layer described before. In the vanilla masked autoencoder, the final layer of the decoder is usually implemented as a fully-connected layer to convert features from latent space into pixel space. For the dynamic decoder layer, we follow the same process used in the dynamic patch embedding to generate the dynamic weights. The only difference is that a fully-connected layer is used in the decoder rather than the convolution layer in the patch embedding layer.

\subsection{Multimodal continual pretraining}

The self-supervised loss formulation is pivotal for training our multimodal EO foundation model. The model leverages the MIM paradigm to avoid the requirement for spatially aligned multimodal datasets. To reduce the computational cost of training on extensive datasets, we propose a continual pretraining strategy in the multimodal setting, incorporating a distillation loss inspired by GFM~\cite{mendieta2023towards} and a weight initialization strategy. 

Considering the varying number of channels in different EO modalities, directly using ImageNet pretrained weights for continual pretraining is impossible. Instead, we design a proxy-based distillation method that extracts optical data as a proxy to ensure representation similarity between the teacher and student networks. As illustrated in Fig.~\ref{DPE} (2), for multi-channel input data with more than three channels, we extract the RGB channels to form a three-channel input $\mathbf{X}_{p} \in \mathbb{R}^{3\times H\times W}$. We randomly select one channel for Sentinel-1 data with only two bands and duplicate it to a synthetic three-channel image. This input $\mathbf{X}_{p}$ is then fed into an ImageNet-pretrained teacher model to get teacher features $\mathbf{F}_{t}$. Concurrently, the dynamic encoder in DOFA is also used to encode $\mathbf{X}_{p}$ into student features $\mathbf{F}_{s}$. Throughout this procedure, the teacher model's weights remain frozen to preserve structured representations and reduce the computational load during optimization. 

We also follow a continual pretraining strategy for initializing the dynamic weight generator. First, we pretrain the weight generator to mimic the teacher model's patch embedding layer weights. We then use the pretrained weights to initialize the dynamic embedding layer. The training loss comprises two distinct components. One is the MIM reconstruction loss, which forces the model to predict $\mathbf{X'} \in \mathbb{R}^{C\times H\times W}$ for reconstructing various data modalities from the full-channel inputs $\mathbf{X}$. The other is the feature distillation loss, which employs the cosine similarity between the teacher and student feature representations to guide the student model. Suppose that the encoded feature of the full channel input is $\mathbf{F}$; then the composite loss function can be formulated as follows:
\begin{equation}
\mathcal{L} = \frac{1}{N} \sum_{i=1}^{N} \| \mathbf{X}_{i} - \mathbf{X'}_{i} \|^2 - \frac{\mathcal{F}_P(\mathbf{F}_{s}^i) \cdot \mathbf{F}_{t}^i}{\|\mathcal{F}_{P}(\mathbf{F}_{s}^i)\|_2 \cdot \|\mathbf{F}_{t}^i\|_2},
\end{equation}
where $\mathcal{F}_P$ is a linear projection layer, and $N$ is the number of data samples. 

The model training is supervised from two distinct perspectives. First, it leverages the complete spectral information present in the input to learn cross-modal features via image reconstruction. Second, it distills knowledge from extensively pretrained models into diverse data modalities using a single dynamic model. This approach enables efficient and robust feature learning across various modalities.

\subsection{DOFA+: Hybrid Continual Pre-training}
To further enhance both performance and efficiency, we extend DOFA with three key improvements:

\textbf{Adaptation from Strong Priors} Starting from DINOv2~\cite{oquab2023dinov2} weights, we apply the MIM objective to enable efficient adaptation to multiple EO modalities. This preserves the strong semantic priors of DINOv2 while guiding the model to learn EO-specific visual patterns through local reconstruction.

\textbf{Hierarchical Feature Distillation} To complement the MIM objective, we introduce a hierarchical distillation strategy that aligns the student's intermediate representations with those of the DINOv2 teacher across multiple layers. 

\textbf{Compact Pre‑training Regime} We demonstrate that high‑quality adaptation does not require billion‑scale corpora: {DOFA+} attains state‑of‑the‑art performance after seeing only 410k EO image tiles, much fewer samples to reduce both cost and carbon footprint.

Building on these refinements, we present {DOFA+}, a lightweight, universal Earth‑observation encoder that combines the reconstruction pressure of MAE with the semantic guidance of a vision foundation model. During training, the student reconstructs randomly masked patches while simultaneously distilling hierarchical supervision signals from \textsc{DINOv2}. This recipe offers two practical benefits:
\begin{enumerate}
\item \textbf{Simplicity} DOFA+ maintains a single-branch architecture with a minimal training objective, avoiding the need for complex multi-task losses or auxiliary network modules.
\item \textbf{Complementary Learning} DOFA+ combines local detail learning and global semantic alignment in a lightweight continual pretraining setup, enabling efficient adaptation to diverse EO tasks without retraining the model.
\end{enumerate}

\noindent The training objective of DOFA+ combines a reconstruction loss from masked image modeling and a multi-level feature distillation loss based on cosine similarity. Let $\mathbf{X} \in \mathbb{R}^{C \times H \times W}$ denote the full-spectrum input, and $\mathbf{X}'$ the reconstructed output. Let ${\mathbf{F}{s}^{i,l}}$ and ${\mathbf{F}{t}^{i,l}}$ represent the student and teacher features from the $l$-th layer, respectively, for sample $i$.

The total training loss is defined as:
\begin{equation}
\small
\mathcal{L} = \underbrace{\frac{1}{N} \sum_{i=1}^{N} \| \mathbf{X}_{i} - \mathbf{X'}_{i} \|^2}_{\text{MIM Reconstruction Loss}} 
- \lambda \cdot \underbrace{\frac{1}{L} \sum_{l=1}^{L} \frac{\mathcal{F}_P^l(\mathbf{F}_{s}^{i,l}) \cdot \mathbf{F}_{t}^{i,l}}{\|\mathcal{F}_P^l(\mathbf{F}_{s}^{i,l})\|_2 \cdot \|\mathbf{F}_{t}^{i,l}\|_2}}_{\text{Multi-layer Feature Distillation}}
\label{eq:dofa_plus_loss}
\end{equation}
\noindent Here, $\mathcal{F}_P^l(\cdot)$ is a linear projection aligning student features to the teacher's dimension, $L$ is the number of distillation layers, $N$ is the batch size, and $\lambda$ is a balancing coefficient.

This training procedure enables DOFA+ to learn semantically rich, transferable representations, supporting scalable and general-purpose EO foundation models with strong cross-modal generalization and low resource demands.


\section{Experiments}

We conduct extensive experiments across 22 datasets encompassing various Earth Observation (EO) tasks, including image classification, semantic segmentation, object detection, and change detection. We assess model performance under several settings: linear probing, decoder fine-tuning, and full fine-tuning. To demonstrate generalizability, we use diverse datasets with varying resolutions, spectral modalities, and geospatial regions. We emphasize energy-efficient training protocols by limiting training epochs and freezing pre-trained backbones wherever possible.

\begin{table*}[htbp]
\newcommand\dd{0}
\centering
\caption{\textbf{Linear probing results on six classification tasks.} All models are trained for 50 epochs. The reported numbers are top-1 overall accuracy (OA). The m-bigearthnet dataset is evaluated using micro-averaged multilabel average precision. Missing values are due to the inability of the model to adapt to this domain.}
\label{cls_res}
\fontsize{7.5pt}{7.5pt}\selectfont
\begin{tabular}{lccccccc}
\toprule
    \rotatebox[origin=c]{\dd}{Method}          &\rotatebox[origin=c]{\dd}{Backbone}            & \rotatebox[origin=c]{\dd}{m-bigearthnet}  & \rotatebox[origin=c]{\dd}{m-forestnet}    & \rotatebox[origin=c]{\dd}{m-brick-kiln}   & \rotatebox[origin=c]{\dd}{m-pv4ger}       & \rotatebox[origin=c]{\dd}{m-so2sat}       & \rotatebox[origin=c]{\dd}{m-eurosat} \\ \midrule
MAE\_Single~\cite{he2022masked}          & ViT-B                  & 63.6          & -          & 88.9          & 92.2          & 50.0          & 89.0          \\
OFA-Net~\cite{OFA}  & ViT-B             & 65.0          & - & 94.7          & 93.2          & 49.4          & 91.9          \\ 
SatMAE~\cite{cong2022satmae} & ViT-B         & 62.1          & -             & 93.9          & -             & 46.9          & 86.4          \\
Scale-MAE~\cite{reed2023scale}    & ViT-L   & -       & -             & -             & 96.9          & -             & -             \\
GFM~\cite{mendieta2023towards}    & Swin-B  &-  &-  &-  &96.8  &-  &- \\
Cross-Scale MAE~\cite{tang2024cross} &ViT-B  &-  &-  &-  &93.1  &-  &- \\
FG-MAE~\cite{wang2023feature}       & ViT-B                  & 63.0          & -             & 94.7          & -             & 51.4          & 87.0          \\
CROMA~\cite{fuller2023croma} &ViT-B                & {67.4} & -             & 91.0          & -             & 49.2          & 90.1          \\
AnySat~\cite{astruc2025anysat} &ViT-B                & {63.8} & 50.9             & 90.3          & 92.8             & 42.5          & 87.6          \\
Galileo~\cite{tseng2025galileo} &ViT-B                & {65.1} & -             & 93.1          & -             & 54.2          & 88.6          \\
\midrule
\textbf{DOFA} &ViT-B               & {65.7}         & 50.9         & 95.8         & 96.9         & 55.1         & 93.9         \\
\textbf{DOFA} &ViT-L               & \underline{67.5}         & \underline{54.6}          & \textbf{96.9} & \underline{97.3} & \underline{60.1} & \textbf{97.1} \\ 
\midrule
\textbf{DOFA+} &ViT-B               & 64.8         & 49.8         & 94.9        & 96.9         & {57.5}        & 93.0        \\
\textbf{DOFA+} &ViT-L               & \textbf{68.3}         & \textbf{55.1}          & \textbf{96.9} & \textbf{97.9} & \textbf{61.3} & \underline{95.3} \\  
\bottomrule
\end{tabular}
\end{table*}

\subsection{GEO-Bench Experiments}

\subsubsection{GEO-Bench Classification Experiments}
There are six image classification datasets provided in GEO-Bench~\cite{lacoste2023geo}: m-bigearthnet, m-so2sat, m-brick-kiln, m-forestnet, m-eurosat, and m-pv4ger. These datasets span diverse domains and applications, including forest monitoring, land use classification, and infrastructure detection. They are sourced from multiple satellite platforms, covering different spectral ranges and spatial resolutions.

On the classification datasets, we follow the common practice of using RandomResizedCrop (scale 0.8 to 1.0) and RandomHorizontalFlip as data augmentations. The default crop size is 224$\times$224 for all datasets and baseline models except SatMAE~\cite{cong2022satmae} and CROMA~\cite{fuller2023croma}, of which the crop size is 96$\times$96 for SatMAE and 120$\times$120 for CROMA, following the official setup to match their smaller patch size of 8. We optimize cross-entropy loss for most datasets, except for m-bigearthnet, for which the multi-label soft margin loss is used. The LARS optimizer is utilized with cosine decay to train the last linear layer of each foundation model for 50 epochs. Considering the wide range of diversity among existing foundation models, we employ dataset-specific learning rates and batch sizes tailored to enhance the performance of classification tasks. We sweep over a grid search to pick the best learning rate from [0.5, 1.0, 10, 20] for each dataset. 

Table~\ref{cls_res} presents the classification results. Our proposed DOFA and DOFA+ demonstrate superior performance across multiple datasets, validating their strong generalization capability with minimal training cost. Regarding the flexibility, DOFA and DOFA+ can be adapted to diverse EO modalities without changing the model architectures or retraining any part of the model. Unlike other models that often require modality-specific adaptation, DOFA adapts seamlessly to unseen sensors. For instance, DOFA achieves strong performance on m-forestnet, despite never seeing the Landsat-8 data during pretraining. This cross-modal adaptability verifies DOFA’s utility as a general-purpose EO foundation model.

\begin{table*}[htbp]
\newcommand\dd{0}
\centering
\caption{\textbf{Partial fine-tuning results on six segmentation tasks.} All models are trained with a frozen backbone for 20 epochs. Reported numbers are mean intersection over union (mIoU). Missing values are due to the inability of the model to readily adapt to this domain.}\label{seg_res}
\fontsize{7.5pt}{7.5pt}\selectfont
\begin{tabular}{lccccccc}
\toprule
\rotatebox[origin=c]{\dd}{Method} &\rotatebox[origin=c]{\dd}{Backbone} &\rotatebox[origin=c]{\dd}{m-pv4ger-seg} &\rotatebox[origin=c]{\dd}{m-nz-cattle} &\rotatebox[origin=c]{\dd}{m-NeonTree} &\rotatebox[origin=c]{\dd}{m-cashew-plant} &\rotatebox[origin=c]{\dd}{m-SA-crop} &\rotatebox[origin=c]{\dd}{m-chesapeake} \\ \midrule
MAE\_Single~\cite{he2022masked} & ViT-B &88.4 &76.4 &53.0 &40.7 &30.7 &51.9 \\
OFA-Net~\cite{OFA} & ViT-B &89.4 &77.6 &53.3 &47.9 &31.9 &54.5 \\
Scale-MAE~\cite{reed2023scale} & ViT-L &83.5 &76.5 &51.0 &- &- &61.0 \\
GFM~\cite{mendieta2023towards} & Swin-B &92.0 &75.0 &51.1 &- &- &63.8 \\
Cross-Scale MAE~\cite{tang2024cross} & ViT-B &83.2 &77.9 &52.1 &- &- &52.3 \\
CROMA~\cite{fuller2023croma}& ViT-B &- &- &- &30.1 &31.4 &- \\
FG-MAE~\cite{wang2023feature} & ViT-B &- &- &- &40.8 &30.6 &- \\
AnySat~\cite{astruc2025anysat} &ViT-B                & {89.9} & 76.7     & 52.2          & 27.2             & 26.8          & 54.0          \\
\midrule
\textbf{DOFA} &ViT-B &{94.5} &{81.4} &{58.8} &51.5 &\underline{33.0} &{65.3} \\
\textbf{DOFA} &ViT-L &{95.0} &{81.8} &{59.4} &\textbf{56.9} &{32.1} &\textbf{66.3} \\
\midrule
\textbf{DOFA+} &ViT-B &\underline{95.1} &\underline{82.7} &\underline{61.1} &\underline{55.1} &{30.2} &{70.8} \\
\textbf{DOFA+} &ViT-L &\textbf{95.7} &\textbf{83.5} &\textbf{63.8} &\textbf{60.2} &\textbf{33.2} &\textbf{71.6} \\ \bottomrule
\end{tabular}
\end{table*}

\subsubsection{GEO-Bench Segmentation Experiments}
The six segmentation datasets include m-pv4ger-seg, m-chesapeake-landcover, m-cashew-plantation, m-SA-crop-type, m-nz-cattle, and m-NeonTree. These datasets cover tasks like solar panel mapping, land cover segmentation, crop classification, and canopy delineation, across RGB, multispectral, and hyperspectral modalities. We freeze the encoder and train a UPerNet~\cite{xiao2018unified} decoder for segmentation tasks. For all the models except the GFM with Swin Transformer, we transform the features into a feature pyramid with channels 512 and four different scales: 4, 2, 1, 0.5. The UPerNet segmentation head is then used to output the segmentation results. We use the AdamW optimizer, batch size 64, and an initial learning rate of 0.005 with cosine decay for 20 epochs for segmentation tasks. The learning rate is relatively stable across datasets for segmentation tasks.

We use center crop, random rotation, and random horizontal and vertical flips for segmentation tasks. Images of each dataset are normalized based on the dataset's mean and standard deviation. Table~\ref{seg_res} shows that DOFA and DOFA+ consistently outperform other foundation models across all datasets. Specifically, DOFA with ViT-Base and ViT-Large backbones achieves particularly high accuracy on both the m-NeonTree and m-nz-cattle datasets, highlighting the strength of its high-capacity architecture. On the segmentation datasets, DOFA+ shows a significant performance improvement compared with DOFA. We attribute this improvement to the multi-layer distillation method, which ensures the model learns fine-grained spatial patterns. These results further confirm DOFA’s robustness, adaptability, and efficiency across segmentation tasks.

\subsection{PANGEA Benchmark Experiments}
We further evaluate DOFA+ on the PANGEA benchmark, which covers a diverse range of EO downstream tasks. In this setting, we perform decoder fine-tuning only, while freezing the pretrained DOFA and DOFA+ backbones. All models are trained for 100 epochs with fixed optimization hyperparameters to ensure fairness. To maintain consistency, we restrict our experiments to the datasets for which the original benchmark provides direct download links, ensuring that comparisons are conducted under the same conditions. Training hyperparameters are kept identical to those specified in the original PANGEA paper. The indices used for combining intermediate layers may be tuned for different datasets. For input resolution, if the original image size is smaller than 224, we upscale it to 224 for DOFA+ since DOFA+ is pretrained with an image size of 224. If it is larger, we retain the original size, following the UNet baseline setting.

Table~\ref{tab:pangea} summarizes the results across eight representative downstream tasks. 
For the results of DOFA, we include the results from the PANGEA~\cite{marsocci2024pangaea} benchmark. DOFA with ViT-Base already achieves competitive performance against existing foundation models, outperforming RemoteCLIP and SatlasNet by a clear margin, and reaching a comparable average mIoU to CROMA with fewer computational costs. More importantly, DOFA exhibits consistent performance across tasks of different types, including burned area mapping (BurnSr), flood mapping (Sen1Floods11), and crop monitoring (AI4Farms). This demonstrates that DOFA effectively learns transferable EO-specific spectral-spatial representations.

When scaling to DOFA+ (ViT-Large), we observe substantial improvements across nearly all tasks, setting new state-of-the-art (SOTA) results on the PANGEA benchmark. Overall, DOFA+ attains the highest average mIoU of \textbf{59.81}, outperforming TerraMindv1-L, despite being trained with substantially fewer pretraining images and significantly lower computational resources. It also achieves the best or second-best scores in 6 out of 8 tasks, with particularly strong performance in BurnSr (\textbf{86.53}), CTM-SS (\textbf{57.47}), and SN7 (\textbf{63.06}). These tasks involve heterogeneous data sources, indicating the robustness of DOFA+ in handling diverse EO modalities.
Although TerraMindv1-L achieves the best score on the MADOS dataset, DOFA+ still delivers the second-best performance. Importantly, unlike the TerraMind models, DOFA+ does not rely on LULC maps during pretraining, even though such maps can provide a strong advantage for Sentinel-2–based segmentation tasks. Overall, DOFA+ attains a superior average rank (\textbf{2.25}) across all benchmarks, highlighting both its strong accuracy and its stability across diverse downstream tasks.

\begin{table*}[htbp]
  \centering
  \caption{Benchmark results across 8 EO downstream tasks.  An asterisk ($^\ast$) marks tasks that use only single-scene inputs.  Best scores in each column are \textbf{bold}; second–best are \underline{underlined}.}
  \setlength{\tabcolsep}{4pt}
  \scalebox{0.93}{
  \begin{tabular}{lcccccccc|c}
    \toprule
    \textbf{Model} & BurnSr$^\ast$ & MADOS$^\ast$ & PASTIS & Sen1Fl11 & DEN$^\ast$ & CTM-SS & SN7$^\ast$ & AI4Farms$^\ast$ & Avg.\ mIoU \\
    \midrule
    CROMA             & 82.42 & 67.55 & 32.32 & \underline{90.89} & 38.29 & 49.38 & 59.28 & 25.65 & 55.72\\
    DOFA (Base)       & 80.63 & 59.58 & 30.02 & 89.37 & \underline{39.29} & 51.33 & 61.84 & 27.07 & 54.89\\
    GFM-Swin          & 76.90 & 64.71 & 21.24 & 72.60 & 34.09 & 46.98 & 60.89 & 27.19 & 50.58 \\
    Prithvi 1.0 100M  & 83.62 & 49.98 & 33.93 & 90.37 & 27.86 & 43.07 & 56.54 & 26.86 & 51.53 \\
    RemoteCLIP        & 76.59 & 60.00 & 18.23 & 74.26 & 31.78 & 52.05 & 57.76 & 25.12 & 49.47 \\
    SatlasNet         & 79.96 & 55.86 & 17.51 & 90.30 & 36.31 & 46.97 & 61.88 & 25.13 & 51.74 \\
    Scale-MAE         & 76.68 & 57.32 & 24.55 & 74.13 & 35.11 & 25.42 & \underline{62.96} & 21.47 & 47.20 \\
    SpectralGPT       & 80.47 & 57.99 & 35.44 & 89.07 & 37.85 & 46.95 & 58.86 & 26.75 & 54.17\\
    S.\textsubscript{-S12}-MoCo      & 81.58 & 51.76 & 34.49 & 89.26 & 35.44 & 48.58 & 57.64 & 25.38 & 53.02 \\
    S.\textsubscript{-S12}-DINO      & 81.72 & 49.37 & 36.18 & 88.61 & 34.81 & 48.66 & 56.47 & 25.62 & 52.68 \\
    S.\textsubscript{-S12}-MAE       & 81.91 & 49.90 & 32.03 & 87.79 & 34.08 & 45.80 & 57.13 & 24.69 & 51.67 \\
    S.\textsubscript{-S12}-Data2Vec  & 81.91 & 44.36 & 34.32 & 88.15 & 35.90 & 54.03 & 58.23 & 24.23 & 52.64  \\
    SMARTIES-B & 82.80 & --- & --- & --- & 38.50 & --- & 62.20 & --- & --- \\
    \midrule
    UNet Baseline     & \underline{84.51} & 54.79 & 31.60 & \textbf{91.42} & {39.46} & 47.57 & 62.09 & \textbf{46.34} & 57.22 \\
    ViT Baseline      & 81.58 & 48.19 & 38.53 & 87.66 & 36.83 & 44.08 & 52.57 & 38.37 & 53.48 \\
    \midrule
    TerraMindv1-B-single & 84.00 & 65.01 & \underline{40.80} & 90.32 & --    & 52.66 & 59.71 & 27.71 & 52.53  \\
    TerraMindv1-B        & 82.42 & {69.52} & 40.51 & 90.62 & 37.87 & {55.80} & 60.61 & 28.12 & 58.18 \\
    TerraMindv1-L        & 82.93 & \textbf{75.57} & \textbf{43.13} & 90.78 & 37.89 & \underline{55.04} & 59.98 & 27.47 & \underline{59.10} \\ \midrule
    DOFA+ (Large)               & \textbf{86.53}    & \underline{69.77}    & 40.26    & 90.38    & \textbf{41.12}    & \textbf{57.47}    & \textbf{63.06}    & \underline{29.93}    & \textbf{59.81} \\
    \bottomrule
  \end{tabular}}
  \label{tab:pangea}
\end{table*}

\subsubsection{Image Classification on RESISC45}
In addition to the datasets in GEO-Bench, we compare DOFA and DOFA+ with existing foundation models on the widely-used RESISC-45 dataset~\cite{cheng2017remote}, which contains 31,500 remote sensing images across 45 scene categories. Table~\ref{resic45} presents the classification results under both frozen backbone evaluation (linear probing) and full finetuning. With the ViT-Base backbone, DOFA achieves 91.3\% (frozen) and 97.3\% (finetuned), already surpassing existing methods. When scaling to ViT-Large, DOFA reaches 91.9\% (frozen) and 97.8\% (finetuned), outperforming other models in both settings.

The performance gains become even more pronounced with DOFA+. Using ViT-Base, DOFA+ improves frozen accuracy to 93.7\%, while finetuning yields 97.5\%. With ViT-Large, DOFA+ achieves the best overall results of \textbf{95.3\%} (frozen) and \textbf{98.1\%} (finetuned), significantly outperforming all prior methods. For the RESISC45 experiments, we apply global mean pooling to the penultimate layer of DOFA and DOFA+, and use the pooled features as input to a linear classification layer. We choose the penultimate layer rather than the final layer because the latter is more directly influenced by the reconstruction objective during pretraining, whereas the penultimate representations tend to capture more generalizable semantic features, leading to better transfer performance on classification tasks.

\begin{table}[htbp]
\centering
\caption{Classification results on the RESISC-45 dataset. The best results are shown in bold.}
\label{resic45}
\scalebox{0.82}{
\begin{tabular}{lcccc}
\toprule
\textbf{Methods} & \textbf{Venue} &\textbf{Backbone} & \textbf{Frozen} & \textbf{Finetune} \\
\midrule
SatMAE~\cite{cong2022satmae} &NeurIPS 2022 & ViT-L & 88.3 & 94.8 \\
ConvMAE~\cite{gao2022convmae} &NeurIPS 2022 & ConvVit-L & 81.2 & 95.0 \\
Scale-MAE~\cite{reed2023scale} &ICCV 2023 & ViT-L & 89.6 & 95.7 \\
Vanilla MAE~\cite{he2022masked} &CVPR 2022 & ViT-L & 88.9 & 93.3 \\
SMARTIES~\cite{sumbul2025smarties} &ICCV 2025 &ViT-L &-- & 95.8 \\
SatMAE++~\cite{noman2024rethinking} &CVPR 2024 &ViT-L &-- & 97.5 \\
\midrule
\textbf{DOFA} &-- &ViT-B &{91.3} & {97.3} \\
\textbf{DOFA} &-- & ViT-L &{91.9} & \underline{97.8} \\ 
\midrule
\textbf{DOFA+} &-- &ViT-B &\underline{93.7} & {97.5} \\
\textbf{DOFA+} &-- & ViT-L &\textbf{95.3} & \textbf{98.1} \\ \bottomrule
\end{tabular}}
\end{table}

\subsection{Object Detection Experiments}
We evaluate object detection performance on the DIOR dataset using the Faster R-CNN detector with DOFA and DOFA+ backbones. The DIOR dataset is a large-scale benchmark for remote sensing object detection, containing 23,463 images and 192,472 annotated instances across 20 object categories, covering diverse scenes with significant variation in scale, orientation, and background complexity. Following the existing experimental setting, Faster R-CNN is adopted as the detection head, and all models are fully finetuned. For DOFA and DOFA+, we train the models for 15 epochs with a learning rate of 1e-4 and batch size 16. For the evaluation metric, $mAP_{50}$ is used, which refers to the mean Average Precision computed at an Intersection-over-Union threshold of 0.5, which is a standard metric for evaluating object detection performance.

Table~\ref{tab:dior_horizontal} reports the detection performance in terms of mAP$_{50}$. DOFA already achieves competitive results with an mAP$_{50}$ of 76.21, while DOFA+ establishes a new SOTA with \textbf{79.73}, surpassing strong baselines such as RingMo (75.90), CMID (75.11), SkySense (78.73), and even the high-capacity SkySense V2 (79.50). Note that, compared with SkySense V2, our model uses ViT-L, which is considerably more computationally efficient than Swin-Huge. Furthermore, our pretraining dataset and GPU hours are significantly smaller, highlighting the efficiency of our approach. 

Unlike SkySense models, which use separate backbones for optical RGB and multispectral data, DOFA and DOFA+ employ a single set of model parameters across all tasks. Despite this unified design, they adapt effectively to the high-resolution characteristics of the DIOR dataset, while also delivering strong performance on Sentinel-1 and Sentinel-2 benchmarks that involve much lower-resolution imagery. This highlights the flexibility of DOFA+ in handling diverse input conditions, including different sensors, GSD, and spectral band combinations. The ability to seamlessly adapt to varying data modalities while maintaining state-of-the-art performance underscores the robustness and universality of the DOFA framework.

From these experiments, we conclude that DOFA+ provides both flexibility and scalability: it can effectively exploit spectral-spatial information across datasets with widely differing properties, and it achieves superior performance on high-resolution detection tasks without requiring task-specific architectural modifications.

\begin{table}
  \centering
  \caption{Detection performance on the DIOR horizontal dataset. Faster R-CNN is used as the object detector.}
  \label{tab:dior_horizontal}
  \setlength{\tabcolsep}{7pt}  
  \begin{tabular}{@{} l c c @{}}
    \toprule
    Models & Backbone & DIOR ($\mathbf{mAP_{50}}$) \\
    \midrule
    GASSL~\cite{ayush2021geography}    &ViT-B       & 67.40 \\
    SatMAE~\cite{cong2022satmae}  &ViT-B      & 70.89 \\
    RingMo~\cite{yao2023ringmo}  &ViT-B      & 75.90 \\
    RVSA~\cite{wang2022advancing}      &ViT-B      & 73.22 \\
    BFM~\cite{cha2023billion}        &ViT-B      & --    \\
    TOV~\cite{tao2023tov}        &ViT-B      & 70.16 \\
    SSL4EO~\cite{wang2022ssl4eo}  &ViT-B      & 64.82 \\
    CMID~\cite{muhtar2023cmid}      &ViT-B      & 75.11 \\
    CACo~\cite{mall2023change}      &ViT-B     & 66.91 \\
    SatLas~\cite{bastani2023satlaspretrain}  &ViT-B      & 74.10 \\
    GFM~\cite{mendieta2023towards}        &ViT-B      & 72.84 \\
    Scale-MAE~\cite{reed2023scale} &ViT-B  & 73.81 \\
    MA3E~\cite{li2024masked}    &ViT-B        & --    \\
    SkySense~\cite{guo2023skysense}  &ViT-B  & 78.73 \\
    SkySense V2~\cite{zhang2025skysensev2}    &Swin-H    & \underline{79.50} \\
    \midrule
    DOFA                   & ViT-L   & 76.21 \\
    DOFA+                  & ViT-L   & \textbf{79.73} \\
    \bottomrule
  \end{tabular}
\end{table}

\subsection{Ablation Studies}
We conduct ablation studies to investigate the contributions of key components in DOFA’s pretraining pipeline and to better understand the robustness of the learned representations. Unless otherwise specified, experiments are performed with DOFA (ViT-Base) and DOFA+ (ViT-Base).  

First, we examine the impact of the number of spectral bands used during pretraining. A central question is whether the pretrained DOFA+ encoder can effectively learn and preserve representations for spectral modalities beyond RGB. To this end, we evaluate DOFA+ (ViT-Base) on the Sen1Floods11 dataset using Sentinel-2 imagery with 13 spectral bands. This experiment provides insights into the model’s capacity to generalize across different spectral combinations and its ability to leverage richer spectral information for downstream tasks.  

Second, we analyze the influence of pretraining duration. Specifically, we vary the number of pretraining epochs and evaluate the resulting representations via linear probing on the RESISC45 dataset. This allows us to assess how the length of pretraining affects representation quality and transfer performance, and whether longer pretraining yields diminishing or consistent returns.  

Together, these ablations shed light on how spectral diversity and pretraining duration contribute to DOFA’s effectiveness, providing guidance for scaling the model to different data modalities and training regimes.

\begin{table*}[ht]
\centering
\caption{Ablation Studies on the performance of DOFA+ (ViT-B) using different numbers of spectral channels. Comparison of band combinations is conducted on the Sen1Floods11 dataset.}
\label{tab::abl}
\scalebox{0.84}{
\begin{tabular}{lccc}
\midrule
\textbf{Bands} & \textbf{F1} & \textbf{mIoU} & \textbf{Water IoU} \\
\midrule
RGB & 81.03 & 70.66 & 49.77 \\
\midrule 
RGB + Coastal Aerosol (CA) & 80.48 & 70.02 & 48.63 \\
RGB + CA + RED\_EDGE\_1 (RE1) & 84.71 & 75.27 & 57.51 \\
RGB + CA + RE1 + RE2 & 86.49 & 77.67 & 61.51 \\
RGB + CA + RE1 + RE2 + RE3 & 87.45 & 78.98 & 63.77 \\
RGB + CA + RE1 + RE2 + RE3 + NIR Broad & 88.83 & 80.90 & 67.49 \\
RGB + CA + RE1 + RE2 + RE3 + NIR Broad + NIR Narrow & 89.31 & 81.60 & 68.67 \\
RGB + CA + RE1 + RE2 + RE3 + NIR Broad + NIR Narrow + Water Vapor & 90.12 & 82.84 & 70.58 \\
RGB + CA + RE1 + RE2 + RE3 + NIR Broad + NIR Narrow + Water Vapor + Cirrus & 89.62 & 82.09 & 69.38 \\
RGB + CA + RE1 + RE2 + RE3 + NIR Broad + NIR Narrow + Water Vapor + Cirrus + SWIR1 & 91.85 & 85.53 & 75.03 \\
RGB + CA + RE1 + RE2 + RE3 + NIR Broad + NIR Narrow + Water Vapor + Cirrus + SWIR1 + SWIR2 & \textbf{92.51} & \textbf{86.58} & \textbf{76.83} \\
\bottomrule
\end{tabular}}
\end{table*}

\subsubsection{Number of Spectral Bands}
\begin{figure*}[htbp]
    \centering
    \includegraphics[width=0.92\textwidth]{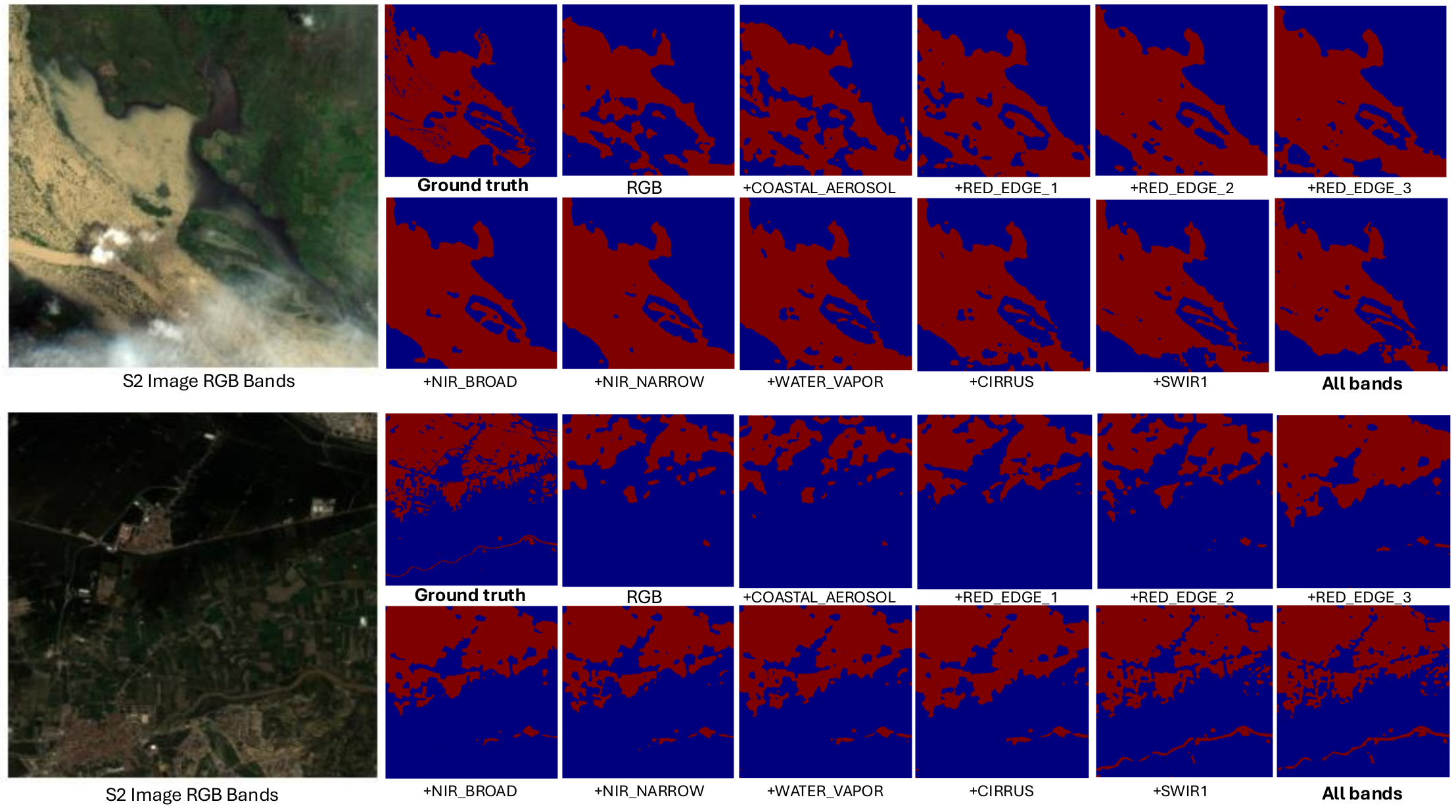}
    \caption{Segmentation maps with progressively added spectral bands (from RGB to full set). As more bands are included, boundaries become sharper and errors decrease, with F1, mIoU, and Water IoU consistently improving and reaching their peak in the full-band configuration.}
    \label{vis:flood}
\end{figure*}

On the Sen1Floods11 dataset, we evaluate DOFA+ (ViT-B) with different combinations of spectral bands to assess how spectral diversity influences segmentation performance. The results are presented in Table~\ref{tab::abl}. It is worth noting that in this ablation study, we do not tune hyperparameters, but focus solely on analyzing the effect of different spectral bands. Starting from the RGB setting, the model achieves an F1 score of 81.03. As additional bands are incrementally introduced, performance consistently improves across all metrics, indicating that the pretrained DOFA+ backbone is capable of effectively exploiting richer spectral information.
\begin{figure}[htbp]
    \centering
    \includegraphics[width=0.475\textwidth]{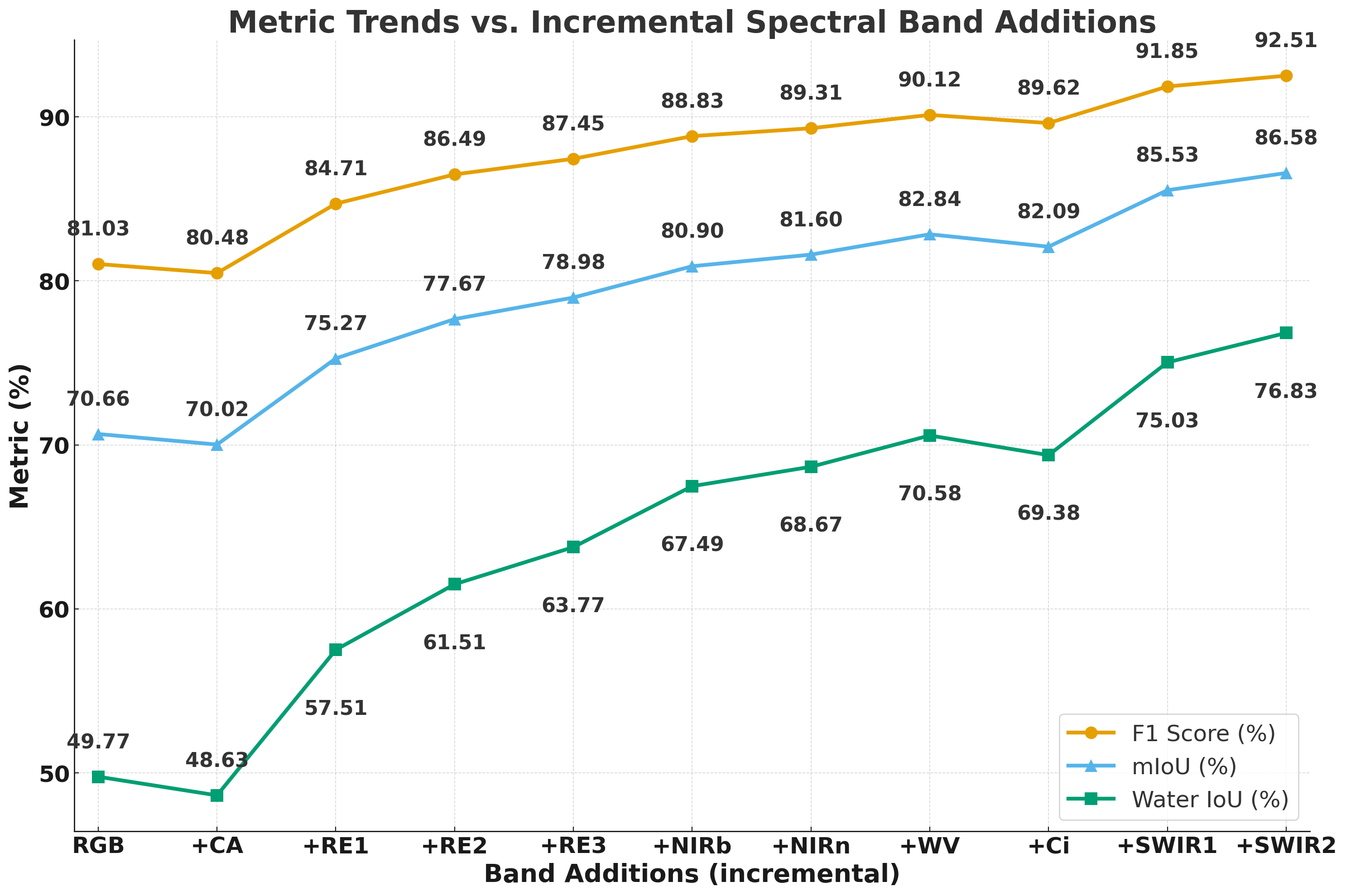}
    \caption{Trend of segmentation performance as additional spectral bands are incrementally included. Starting from RGB, performance improves significantly with the addition of more bands.}
    \label{fig::trend}
\end{figure}
Adding the Coastal Aerosol (CA) band doesn't improve the performance compared to RGB, but subsequent inclusion of Red Edge (RE) bands leads to substantial improvements. With RGB + CA + RE1, the Water IoU increases from 49.77 to 57.51, demonstrating the importance of vegetation-sensitive bands for flood mapping. As shown in Fig.~\ref{fig::trend}, performance continues to rise steadily as more RE and NIR bands are introduced, with the model benefiting from complementary spectral responses across vegetation, soil, and water surfaces. For an intuitive visualization, Fig.~\ref{vis:flood} shows that the segmentation maps become progressively finer and more accurate as additional spectral bands are incorporated.

The most significant gains are observed when SWIR bands are included. With the full combination of spectral bands, DOFA+ achieves the best overall performance, with an F1 score of 92.51, mIoU of 86.58, and Water IoU of 76.83. These results confirm that the pretrained backbone is not only robust to varying spectral inputs but also capable of leveraging additional modalities without requiring architecture changes or retraining the model. Overall, the ablation results validate that DOFA+ learns powerful and flexible spectral–spatial representations during pretraining. Even with a frozen backbone, it adapts effectively to different band combinations and consistently benefits from the inclusion of additional spectral information. 

\begin{table}[t]
  \centering
  \caption{Ablation Studies on the performance of DOFA with different training epochs.}
  \label{tab:epochs}
  \setlength{\tabcolsep}{7pt}  
  \begin{tabular}{@{} c c c @{}}
    \toprule
    Epochs & Backbone & RESISC-45 ($\mathbf{Acc@1}$) \\
    \midrule
    E20  &ViT-B      & 86.22 \\
    E40      &ViT-B  & 87.41 \\
    E60 &ViT-B      & 89.36 \\
    E80 &ViT-B      & 91.12 \\
    \midrule
    E100    & ViT-B   & 91.30 \\
    \bottomrule
  \end{tabular}
\end{table}

\subsubsection{Training Epochs}
The results in Table~\ref{tab:epochs} show a clear trend of improvement as the number of pretraining epochs increases. Accuracy steadily rises from 20 to 80 epochs, indicating that longer pretraining enables DOFA to learn richer and more transferable representations. Beyond 80 epochs, the gains become marginal. This suggests that while sufficient pretraining is important for representation quality, excessively long training may yield diminishing returns. Overall, these results confirm the benefit of extended pretraining while also highlighting a tradeoff between efficiency and performance.

\begin{table}[t]
  \centering
  \caption{Pre-training efficiency comparison of recent EO foundation models (mAP column removed). \textit{N/A} indicates \emph{not applicable} due to missing public checkpoints or sensor mismatches that would lead to unfair comparisons.}
  \label{tab:pretrain_efficiency}
  \setlength{\tabcolsep}{6pt} 
  \scalebox{0.68}{
  \begin{tabular}{@{} l l c c c c @{}}
    \toprule
    \multirow{2}{*}{Model} & \multirow{2}{*}{Backbone} & \multirow{2}{*}{PT Epochs} &
    \multicolumn{2}{c}{PT Data Size} & \multirow{2}{*}{RESISC-45} \\
    \cmidrule(lr){4-5}
      &   &   & S2 & RGB & \\
    \midrule
    SatMAE (S2)~\cite{cong2022satmae}                   & ViT-L                     &  50 & 713K  &  --   & N/A  \\
    SatMAE (S2)~\cite{cong2022satmae}                   & ViT-L                     & 200 & 713K  &  --   & N/A  \\
    SatMAE (RGB)~\cite{cong2022satmae}                  & ViT-L                     & 800 &  --   & 364K  & 94.8 \\
    Scale-MAE~\cite{reed2023scale}                   & ViT-L                     & 800 &  --   & 364K  & 95.7 \\
    CROMA~\cite{fuller2023croma}                          & ViT-B$\,(\times2)$        & 300 &   1M  &  --   & N/A  \\
    SpectralGPT~\cite{hong2024spectralgpt}              & ViT-L                     & 200 & 713K  &  --   & N/A  \\
    SpectralGPT$^{+}$~\cite{hong2024spectralgpt}        & ViT-L                     & 300 &   1M  &  --   & N/A  \\
    S2MAE~\cite{li2024s2mae}                          & ViT-L                     & 200 & 713K  &  --   & N/A  \\
    S2MAE$^{*}$~\cite{li2024s2mae}                    & ViT-L                     & 300 &   1M  &  --   & N/A  \\
    SatMAE$^{++}$ (RGB)~\cite{noman2024rethinking}         & ViT-L                     & 800 &  --   & 364K  & 97.5 \\
    SatMAE$^{++}$ (S2)~\cite{noman2024rethinking}          & ViT-L                     &  50 & 713K  &  --   & N/A  \\
    SMARTIES~\cite{sumbul2025smarties}             & ViT-L                     & 300 & 248K  &  60K  & 95.8 \\
    \midrule
    CROMA~\cite{fuller2023croma}                          & ViT-L$\,(\times2)$        & 600 &   1M  &  --   & N/A  \\
    SpectralGPT~\cite{hong2024spectralgpt}              & ViT-H                     & 200 & 713K  &  --   & N/A  \\
    SpectralGPT$^{+}$~\cite{hong2024spectralgpt}        & ViT-H                     & 300 &   1M  &  --   & N/A  \\
    S2MAE~\cite{li2024s2mae}                          & ViT-H                     & 200 & 713K  &  --   & N/A  \\
    S2MAE$^{*}$~\cite{li2024s2mae}                    & ViT-H                     & 300 &   1M  &  --   & N/A  \\
    SkySense~\cite{guo2023skysense}                    & ViT-L+Swin-H & 780 & 21.5M & 21.5M & 96.3$^{*}$ \\
    SkySenseV2~\cite{zhang2025skysensev2}       & ViT-L+Swin-H & -- & 21.5M & 21.5M & 97.2$^{*}$ \\
    \midrule
    DOFA & ViT-L &--  &3.5M &4.5M &\underline{97.8} \\
    DOFA+ & ViT-L &150  &100K &100K &\textbf{98.1} \\
    \bottomrule
  \end{tabular}}
\end{table}
\begin{figure*}[!]
    \centering
    \begin{subfigure}[b]{1.0\textwidth}
        \centering
        \includegraphics[width=1.0\textwidth]{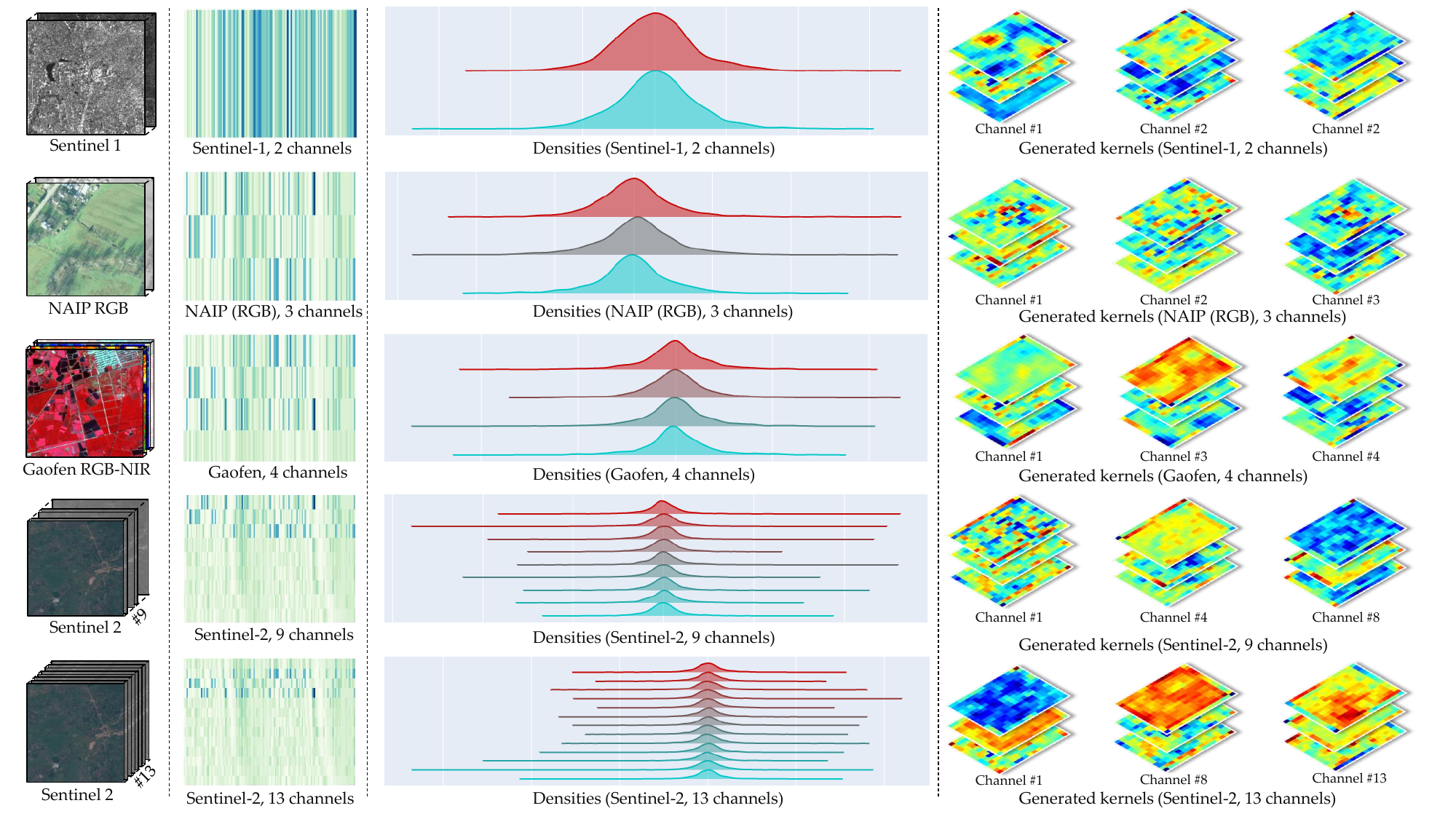}
        \caption{\textbf{Visualization of the dynamic weight generator.} From left to right: examples of input images, learned embeddings for different central wavelengths, the histogram distributions of the generated weights, and some examples of the generated kernel weights.}\label{vis_gweights}
        \captionsetup{font=footnotesize}
    \end{subfigure}
    
    \begin{subfigure}[b]{1.0\textwidth}
        \centering
        \includegraphics[width=1.0\textwidth]{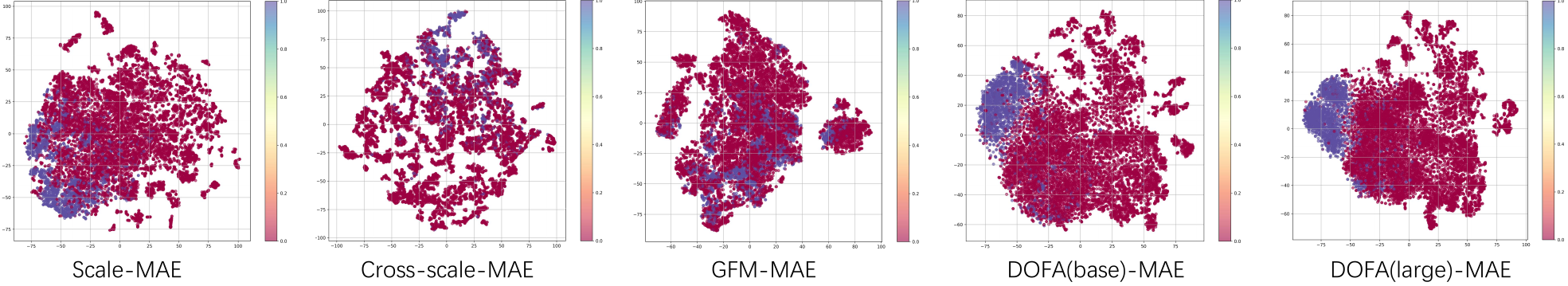}
        \captionsetup{font=footnotesize}
    \end{subfigure}

    \begin{subfigure}[b]{1.0\textwidth}
        \centering
        \includegraphics[width=1.0\textwidth]{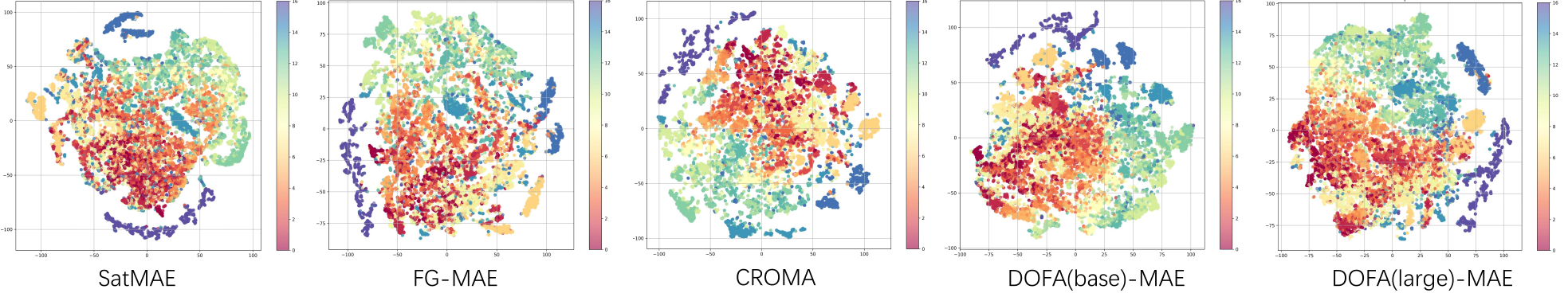}
        \captionsetup{font=footnotesize}
    \end{subfigure}

    \begin{subfigure}[b]{1.0\textwidth}
        \centering
        \includegraphics[width=1.0\textwidth]{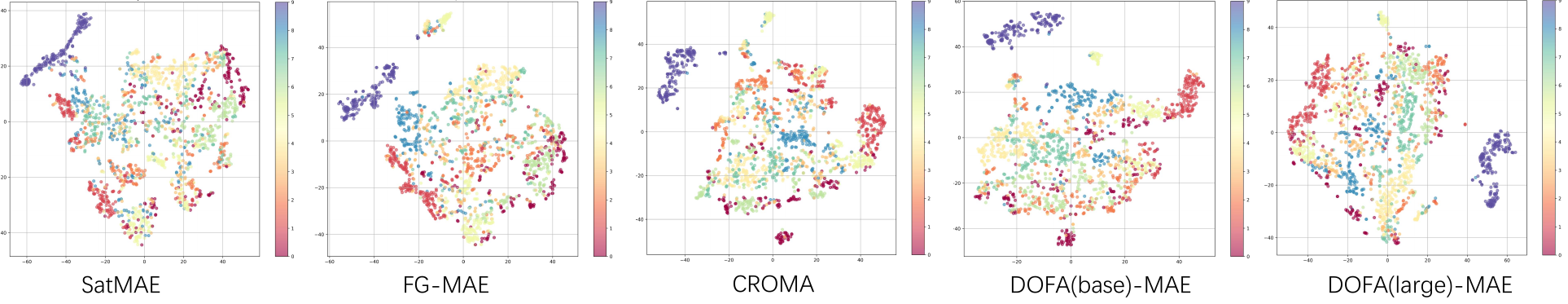}
        \caption{\textbf{t-SNE plots of the feature representations from various foundation models across multiple datasets.} From top to bottom row: the m-pv4ger dataset, the m-so2sat dataset, and the m-eurosat dataset. Enhanced separability signifies more effective representations.} \label{tsne-vis}
        \captionsetup{font=footnotesize}
    \end{subfigure}
    \caption{\textbf{Visualization of learned embeddings.} (a) Visualization of the generated weights for different input modalities. (b) t-SNE plots of the feature representations from various foundation models across multiple datasets.}
    \label{feature_vis}
\end{figure*}
\subsection{Pre-training efficiency comparison}
Table~\ref{tab:pretrain_efficiency} highlights the efficiency advantages of our proposed models, DOFA and DOFA+, in comparison to recent EO foundation models. Unlike prior works that require massive datasets, hundreds of training epochs, and large computational budgets, DOFA and DOFA+ achieve competitive or superior performance with significantly lower resource requirements. 

DOFA is pretrained on a large-scale multimodal EO dataset comprising approximately 11.5 million images from five different modalities: Sentinel-1, Sentinel-2, NAIP, Gaofen, and EnMAP. Pretraining is conducted with a 75\% masking ratio, a batch size of 128, the AdamW optimizer~\cite{loshchilov2017decoupled} (weight decay 0.05), and an initial learning rate of 1.5$e$-4. The learning rate is warmed up for 20 epochs and then decayed using a cosine schedule. The training follows a progressive scheme:  
\begin{itemize}
    \item Initially, the model is trained on a 50K-image subset for 100 epochs.  
    \item Then, it is further trained on a 410K-image subset (100K from each of Sentinel-1, Sentinel-2, NAIP, Gaofen, and 10K from EnMAP) for 20 epochs.  
    \item Finally, we conduct a single epoch of training on the full 11.5M dataset to consolidate representation learning.
\end{itemize}
All images are resized to $224\times224$ and normalized using modality-specific statistics. ViT-Base and ViT-Large teacher models pretrained on ImageNet-21K~\cite{steiner2021train} are used for distillation.

DOFA+ represents a more lightweight variant, designed to validate the efficiency of our distillation strategy under constrained data and compute. It is pretrained for only 150 epochs on a compact dataset of 410K EO images, including 100K samples each from Sentinel-1, Sentinel-2, NAIP, Gaofen, and 10K from EnMAP. All training is conducted using just 8 NVIDIA L40 GPUs (48GB memory each), completing within three days. Despite its lightweight setup, DOFA+ achieves SOTA performance on the RESISC-45 classification benchmark with 98.1\% accuracy. This result is particularly notable when compared to large-scale models such as SatMAE, Scale-MAE, and SatMAE++, which use millions of images and up to 800 training epochs. Training TerraMindv1-B took 12 days on 32 A100 GPUs, totaling 9,216 GPU hours, representing a substantial computational cost. Even more resource-intensive models like SkySenseV2~\cite{zhang2025skysensev2} require 21.5M images, heavy backbones (e.g., Swin-H), and over 44,500 hours of H20 GPU time.

In contrast, DOFA+ employs a single ViT-L backbone with much lower computational cost, yet achieves competitive or even superior performance compared to task-specific counterparts. This demonstrates the strength of our distillation-based pretraining pipeline in extracting rich spectral–spatial representations efficiently, making it suitable for real-world deployment where computational resources are limited.

\subsection{Visualizations}

\textbf{DOFA generates diverse weights dynamically.} We visualize the learned embeddings of various wavelengths and the generated kernels for different sensors in Fig.~\ref{vis_gweights} for a better understanding of DOFA. We randomly select and plot six $16\times 16$ kernel weights for input images with more than four channels. The figures indicate that DOFA can generate weights for different sensors dynamically and effectively.

\textbf{DOFA optimizes separability in latent space.} We visualize the pretrained representations of different models using the dimensionality reduction technique t-SNE~\cite{van2008visualizing} to represent high-dimensional data. Specifically, the extracted features of the pretrained models on downstream datasets m-pv4ger, m-so2sat, and m-eurosat are shown in Fig.~\ref{tsne-vis}. Different colors represent different semantic categories. On these three datasets, the learned features of both versions of DOFA are clustered better than those of other compared models. These figures further validate the effectiveness of the proposed DOFA as a unified EO foundation model.

\section{Conclusion}
In this work, we introduced DOFA and DOFA+, two foundation models for Earth observation (EO) designed to operate flexibly across a wide range of sensors, spectral bands, and spatial resolutions. Unlike prior approaches that are often limited to specific modalities or require excessive compute, our models are built to generalize across tasks and modalities in an efficient and scalable manner. We proposed a wavelength-conditioned dynamic hypernetwork architecture that enables a single model to process multimodal satellite inputs. Through continual and hybrid pretraining across five EO modalities, DOFA learns rich spectral–spatial representations that transfer effectively to downstream tasks. DOFA+ extends this capability further with a lightweight distillation pipeline, offering strong performance even under constrained computational budgets. Comprehensive experiments across multiple benchmarks demonstrate that DOFA and DOFA+ achieve state-of-the-art results. Notably, the models generalize well to unseen sensors and spectral configurations without the need for retraining, highlighting their flexibility and robustness in open-world EO scenarios.

\section{Data availability statements}
In this work, we have constructed an extensive multimodal dataset that is composed of five distinct modalities, each offering unique spectral and spatial data characteristics. In this section, we provide the download links to ensure data availability.

\paragraph{Sentinel-1}
The Sentinel-1 subset of the dataset can be downloaded from \url{https://github.com/allenai/satlas/blob/main/SatlasPretrain.md#download}.
\paragraph{Sentinel-2}
The Sentinel-2 subset of the dataset can be downloaded from \url{https://github.com/allenai/satlas/blob/main/SatlasPretrain.md#download}.
\paragraph{Gaofen}
The Gaofen part of the dataset can be downloaded from \url{https://drive.google.com/drive/folders/1924VnO08Gqo3Nv7Y4KirgJ9kqqCup7f0}.
\paragraph{NAIP}
This NAIP part of the dataset can be downloaded from \url{https://github.com/allenai/satlas/blob/main/SatlasPretrain.md#download}.
\paragraph{EnMAP}
This hyperspectral data from EnMAP used in our dataset can be downloaded from \url{https://hyspecnet.rsim.berlin/}.
\paragraph{Evaluation datasets}
The resisc45 dataset can be downloaded from \url{https://huggingface.co/datasets/timm/resisc45}.

We evaluate the pretrained models on 12 downstream tasks organized in GEO-Bench~\cite{lacoste2023geo}. These datasets cover various applications and data modalities in EO, including six classification tasks and six segmentation tasks. The full guidance for the dataset downloading is available at \url{https://github.com/ServiceNow/geo-bench}. 

All the datasets for evaluation on PANGEA benchmark can be found at \url{https://github.com/VMarsocci/pangaea-bench}.

\section{Code availability}
The training script and inference script have been publicly available at \url{https://github.com/zhu-xlab/DOFA}.
The trained models have been publicly available at \url{https://huggingface.co/earthflow/DOFA}.

\begin{acknowledgements}
The work of Z.X., F.Z., Y.W., F.Z., A.J.S., and X.X.Z is jointly supported by the German Federal Ministry of Education and Research (BMBF) in the framework of the international future AI lab ``AI4EO -- Artificial Intelligence for Earth Observation: Reasoning, Uncertainties, Ethics and Beyond'' (grant number: 01DD20001), by German Federal Ministry for Economic Affairs and Climate Action in the framework of the ``national center of excellence ML4Earth'' (grant number: 50EE2201C), by the German Federal Ministry for the Environment, Nature Conservation, Nuclear Safety and Consumer Protection (BMUV) based on a resolution of the German Bundestag (grant number: 67KI32002B; Acronym: \textit{EKAPEx}) and by Munich Center for Machine Learning. The work of Z. X., I.P., G.C.V., and X. X. Z is also funded by the European Commission through the project ``ThinkingEarth—Copernicus Foundation Models for a Thinking Earth'' under the Horizon 2020 Research and Innovation program (Grant Agreement No. 101130544). 
GCV was partly funded by the European Research Council (ERC) Synergy Grant ``Understanding and Modeling the Earth System with Machine Learning'' (USMILE) under the Horizon 2020 Research and Innovation program (Grant Agreement No. 855187).
\end{acknowledgements}

\bibliographystyle{plainnat}      
\bibliography{template.bib}   

\begin{thebibliography}{55}
\providecommand{\natexlab}[1]{#1}
\providecommand{\url}[1]{\texttt{#1}}
\expandafter\ifx\csname urlstyle\endcsname\relax
  \providecommand{\doi}[1]{doi: #1}\else
  \providecommand{\doi}{doi: \begingroup \urlstyle{rm}\Url}\fi

\bibitem[Agarap(2018)]{agarap2018deep}
Abien~Fred Agarap.
\newblock Deep learning using rectified linear units ({ReLU}).
\newblock \emph{arXiv preprint arXiv:1803.08375}, 2018.

\bibitem[Astruc et~al.(2025)Astruc, Gonthier, Mallet, and Landrieu]{astruc2025anysat}
Guillaume Astruc, Nicolas Gonthier, Cl{\'e}ment Mallet, and Loic Landrieu.
\newblock Anysat: One earth observation model for many resolutions, scales, and modalities.
\newblock In \emph{Proceedings of the IEEE/CVF Conference on Computer Vision and Pattern Recognition (CVPR)}, 2025.

\bibitem[Ayush et~al.(2021)Ayush, Uzkent, Meng, Tanmay, Burke, Lobell, and Ermon]{ayush2021geography}
Kumar Ayush, Burak Uzkent, Chenlin Meng, Kumar Tanmay, Marshall Burke, David Lobell, and Stefano Ermon.
\newblock Geography-aware self-supervised learning.
\newblock In \emph{Proceedings of the IEEE/CVF International Conference on Computer Vision}, pages 10181--10190, 2021.

\bibitem[Bastani et~al.(2023)Bastani, Wolters, Gupta, Ferdinando, and Kembhavi]{bastani2023satlaspretrain}
Favyen Bastani, Piper Wolters, Ritwik Gupta, Joe Ferdinando, and Aniruddha Kembhavi.
\newblock {SatlasPretrain}: A large-scale dataset for remote sensing image understanding.
\newblock In \emph{Proceedings of the IEEE/CVF International Conference on Computer Vision}, pages 16772--16782, 2023.

\bibitem[Camps-Valls et~al.(2021)Camps-Valls, Tuia, Zhu, and Reichstein]{camps2021deep}
Gustau Camps-Valls, Devis Tuia, Xiao~Xiang Zhu, and Markus Reichstein.
\newblock Deep learning for the {Earth} sciences: A comprehensive approach to remote sensing, climate science and geosciences.
\newblock 2021.

\bibitem[Camps-Valls et~al.(2011)Camps-Valls, Tuia, G{\'o}mez-Chova, Jim{\'e}nez, and Malo]{camps2011remote}
Gustavo Camps-Valls, Devis Tuia, Luis G{\'o}mez-Chova, Sandra Jim{\'e}nez, and Jes{\'u}s Malo.
\newblock Remote sensing image processing.
\newblock 2011.

\bibitem[Cepeda et~al.(2023)Cepeda, Nayak, and Shah]{cepeda2023geoclip}
Vicente~Vivanco Cepeda, Gaurav~Kumar Nayak, and Mubarak Shah.
\newblock {GeoCLIP}: Clip-inspired alignment between locations and images for effective worldwide geo-localization.
\newblock \emph{arXiv preprint arXiv:2309.16020}, 2023.

\bibitem[Cha et~al.(2023)Cha, Seo, and Lee]{cha2023billion}
Keumgang Cha, Junghoon Seo, and Taekyung Lee.
\newblock A billion-scale foundation model for remote sensing images.
\newblock \emph{arXiv preprint arXiv:2304.05215}, 2023.

\bibitem[Cheng et~al.(2017)Cheng, Han, and Lu]{cheng2017remote}
Gong Cheng, Junwei Han, and Xiaoqiang Lu.
\newblock Remote sensing image scene classification: Benchmark and state of the art.
\newblock \emph{Proceedings of the IEEE}, 105\penalty0 (10):\penalty0 1865--1883, 2017.

\bibitem[Cong et~al.(2022)Cong, Khanna, Meng, Liu, Rozi, He, Burke, Lobell, and Ermon]{cong2022satmae}
Yezhen Cong, Samar Khanna, Chenlin Meng, Patrick Liu, Erik Rozi, Yutong He, Marshall Burke, David Lobell, and Stefano Ermon.
\newblock {SatMAE}: Pre-training transformers for temporal and multi-spectral satellite imagery.
\newblock \emph{Advances in Neural Information Processing Systems}, 35:\penalty0 197--211, 2022.

\bibitem[Dan and Poo(2004)]{dan2004spike}
Yang Dan and Mu-ming Poo.
\newblock Spike timing-dependent plasticity of neural circuits.
\newblock \emph{Neuron}, 44\penalty0 (1):\penalty0 23--30, 2004.

\bibitem[Dayan and Cohen(2011)]{dayan2011neuroplasticity}
Eran Dayan and Leonardo~G Cohen.
\newblock Neuroplasticity subserving motor skill learning.
\newblock \emph{Neuron}, 72\penalty0 (3):\penalty0 443--454, 2011.

\bibitem[Dosovitskiy et~al.(2020)Dosovitskiy, Beyer, Kolesnikov, Weissenborn, Zhai, Unterthiner, Dehghani, Minderer, Heigold, Gelly, et~al.]{dosovitskiy2020image}
Alexey Dosovitskiy, Lucas Beyer, Alexander Kolesnikov, Dirk Weissenborn, Xiaohua Zhai, Thomas Unterthiner, Mostafa Dehghani, Matthias Minderer, Georg Heigold, Sylvain Gelly, et~al.
\newblock An image is worth 16x16 words: Transformers for image recognition at scale.
\newblock \emph{arXiv preprint arXiv:2010.11929}, 2020.

\bibitem[Fuller et~al.(2023)Fuller, Millard, and Green]{fuller2023croma}
Anthony Fuller, Koreen Millard, and James~R Green.
\newblock {CROMA}: Remote sensing representations with contrastive radar-optical masked autoencoders.
\newblock \emph{arXiv preprint arXiv:2311.00566}, 2023.

\bibitem[Gao et~al.(2022)Gao, Ma, Li, Lin, Dai, and Qiao]{gao2022convmae}
Peng Gao, Teli Ma, Hongsheng Li, Ziyi Lin, Jifeng Dai, and Yu~Qiao.
\newblock Convmae: Masked convolution meets masked autoencoders.
\newblock \emph{arXiv preprint arXiv:2205.03892}, 2022.

\bibitem[Guo et~al.(2023)Guo, Lao, Dang, Zhang, Yu, Ru, Zhong, Huang, Wu, Hu, et~al.]{guo2023skysense}
Xin Guo, Jiangwei Lao, Bo~Dang, Yingying Zhang, Lei Yu, Lixiang Ru, Liheng Zhong, Ziyuan Huang, Kang Wu, Dingxiang Hu, et~al.
\newblock {SkySense}: A multi-modal remote sensing foundation model towards universal interpretation for {Earth} observation imagery.
\newblock \emph{arXiv preprint arXiv:2312.10115}, 2023.

\bibitem[Ha et~al.(2017)Ha, Dai, and Le]{hypernetworks}
David Ha, Andrew~M. Dai, and Quoc~V. Le.
\newblock Hypernetworks.
\newblock In \emph{{ICLR} 2017}, 2017.

\bibitem[He et~al.(2022)He, Chen, Xie, Li, Doll{\'a}r, and Girshick]{he2022masked}
Kaiming He, Xinlei Chen, Saining Xie, Yanghao Li, Piotr Doll{\'a}r, and Ross Girshick.
\newblock Masked autoencoders are scalable vision learners.
\newblock In \emph{Proceedings of the IEEE/CVF Conference on Computer Vision and Pattern Recognition}, pages 16000--16009, 2022.

\bibitem[Hebb(2005)]{hebb2005organization}
Donald~Olding Hebb.
\newblock The organization of behavior: A neuropsychological theory.
\newblock 2005.

\bibitem[Hong et~al.(2024)Hong, Zhang, Li, Li, Li, Yao, Yokoya, Li, Ghamisi, Jia, et~al.]{hong2024spectralgpt}
D~Hong, B~Zhang, X~Li, Y~Li, C~Li, J~Yao, N~Yokoya, H~Li, P~Ghamisi, X~Jia, et~al.
\newblock {SpectralGPT}: Spectral remote sensing foundation model.
\newblock \emph{IEEE Transactions on Pattern Analysis and Machine Intelligence}, 2024.

\bibitem[Irvin et~al.(2023)Irvin, Tao, Zhou, Ma, Nashold, Liu, and Ng]{irvin2023usat}
Jeremy Irvin, Lucas Tao, Joanne Zhou, Yuntao Ma, Langston Nashold, Benjamin Liu, and Andrew~Y Ng.
\newblock {USat}: A unified self-supervised encoder for multi-sensor satellite imagery.
\newblock \emph{arXiv preprint arXiv:2312.02199}, 2023.

\bibitem[Jean et~al.(2019)Jean, Wang, Samar, Azzari, Lobell, and Ermon]{jean2019tile2vec}
Neal Jean, Sherrie Wang, Anshul Samar, George Azzari, David Lobell, and Stefano Ermon.
\newblock {Tile2Vec}: Unsupervised representation learning for spatially distributed data.
\newblock In \emph{Proceedings of the AAAI Conference on Artificial Intelligence}, volume~33, pages 3967--3974, 2019.

\bibitem[Klemmer et~al.(2023)Klemmer, Rolf, Robinson, Mackey, and Ru{\ss}wurm]{klemmer2023satclip}
Konstantin Klemmer, Esther Rolf, Caleb Robinson, Lester Mackey, and Marc Ru{\ss}wurm.
\newblock {SatCLIP}: Global, general-purpose location embeddings with satellite imagery.
\newblock \emph{arXiv preprint arXiv:2311.17179}, 2023.

\bibitem[Lacoste et~al.(2023)Lacoste, Lehmann, Rodriguez, Sherwin, Kerner, L{\"u}tjens, Irvin, Dao, Alemohammad, Drouin, et~al.]{lacoste2023geo}
Alexandre Lacoste, Nils Lehmann, Pau Rodriguez, Evan~David Sherwin, Hannah Kerner, Bj{\"o}rn L{\"u}tjens, Jeremy~Andrew Irvin, David Dao, Hamed Alemohammad, Alexandre Drouin, et~al.
\newblock {GEO-Bench}: Toward foundation models for {Earth} monitoring.
\newblock \emph{arXiv preprint arXiv:2306.03831}, 2023.

\bibitem[Li et~al.(2024{\natexlab{a}})Li, Hong, and Chanussot]{li2024s2mae}
Xuyang Li, Danfeng Hong, and Jocelyn Chanussot.
\newblock S2mae: A spatial-spectral pretraining foundation model for spectral remote sensing data.
\newblock In \emph{Proceedings of the IEEE/CVF Conference on Computer Vision and Pattern Recognition (CVPR)}, 2024{\natexlab{a}}.

\bibitem[Li et~al.(2024{\natexlab{b}})Li, Hou, Ma, Wu, Guo, Ren, and Jiao]{li2024masked}
Zhihao Li, Biao Hou, Siteng Ma, Zitong Wu, Xianpeng Guo, Bo~Ren, and Licheng Jiao.
\newblock Masked angle-aware autoencoder for remote sensing images.
\newblock In \emph{European Conference on Computer Vision}, pages 260--278. Springer, 2024{\natexlab{b}}.

\bibitem[Lillicrap et~al.(2020)Lillicrap, Santoro, Marris, Akerman, and Hinton]{lillicrap2020backpropagation}
Timothy~P Lillicrap, Adam Santoro, Luke Marris, Colin~J Akerman, and Geoffrey Hinton.
\newblock Backpropagation and the brain.
\newblock \emph{Nature Reviews Neuroscience}, 21\penalty0 (6):\penalty0 335--346, 2020.

\bibitem[Loshchilov and Hutter(2017)]{loshchilov2017decoupled}
Ilya Loshchilov and Frank Hutter.
\newblock Decoupled weight decay regularization.
\newblock \emph{arXiv preprint arXiv:1711.05101}, 2017.

\bibitem[Mall et~al.(2023)Mall, Hariharan, and Bala]{mall2023change}
Utkarsh Mall, Bharath Hariharan, and Kavita Bala.
\newblock Change-aware sampling and contrastive learning for satellite images.
\newblock In \emph{Proceedings of the IEEE/CVF Conference on Computer Vision and Pattern Recognition}, pages 5261--5270, 2023.

\bibitem[Manas et~al.(2021)Manas, Lacoste, Gir{\'o}-i Nieto, Vazquez, and Rodriguez]{manas2021seasonal}
Oscar Manas, Alexandre Lacoste, Xavier Gir{\'o}-i Nieto, David Vazquez, and Pau Rodriguez.
\newblock Seasonal contrast: Unsupervised pre-training from uncurated remote sensing data.
\newblock In \emph{Proceedings of the IEEE/CVF International Conference on Computer Vision}, pages 9414--9423, 2021.

\bibitem[Marsocci et~al.(2024)Marsocci, Jia, Bellier, Kerekes, Zeng, Hafner, Gerard, Brune, Yadav, Shibli, et~al.]{marsocci2024pangaea}
Valerio Marsocci, Yuru Jia, Georges~Le Bellier, David Kerekes, Liang Zeng, Sebastian Hafner, Sebastian Gerard, Eric Brune, Ritu Yadav, Ali Shibli, et~al.
\newblock Pangaea: A global and inclusive benchmark for geospatial foundation models.
\newblock \emph{arXiv preprint arXiv:2412.04204}, 2024.

\bibitem[Mendieta et~al.(2023)Mendieta, Han, Shi, Zhu, and Chen]{mendieta2023towards}
Mat{\'\i}as Mendieta, Boran Han, Xingjian Shi, Yi~Zhu, and Chen Chen.
\newblock Towards geospatial foundation models via continual pretraining.
\newblock In \emph{Proceedings of the IEEE/CVF International Conference on Computer Vision}, pages 16806--16816, 2023.

\bibitem[Muhtar et~al.(2023)Muhtar, Zhang, Xiao, Li, and Gu]{muhtar2023cmid}
Dilxat Muhtar, Xueliang Zhang, Pengfeng Xiao, Zhenshi Li, and Feng Gu.
\newblock {CMID}: A unified self-supervised learning framework for remote sensing image understanding.
\newblock \emph{IEEE Transactions on Geoscience and Remote Sensing}, 2023.

\bibitem[Noman et~al.(2024)Noman, Naseer, Cholakkal, Anwer, Khan, and Khan]{noman2024rethinking}
Mubashir Noman, Muzammal Naseer, Hisham Cholakkal, Rao~Muhammad Anwer, Salman Khan, and Fahad~Shahbaz Khan.
\newblock Rethinking transformers pre-training for multi-spectral satellite imagery.
\newblock In \emph{Proceedings of the IEEE/CVF Conference on Computer Vision and Pattern Recognition}, pages 27811--27819, 2024.

\bibitem[Oquab et~al.(2023)Oquab, Darcet, Moutakanni, Vo, Szafraniec, Khalidov, Fernandez, Haziza, Massa, El-Nouby, et~al.]{oquab2023dinov2}
Maxime Oquab, Timoth{\'e}e Darcet, Th{\'e}o Moutakanni, Huy Vo, Marc Szafraniec, Vasil Khalidov, Pierre Fernandez, Daniel Haziza, Francisco Massa, Alaaeldin El-Nouby, et~al.
\newblock Dinov2: Learning robust visual features without supervision.
\newblock \emph{arXiv preprint arXiv:2304.07193}, 2023.

\bibitem[Reed et~al.(2023)Reed, Gupta, Li, Brockman, Funk, Clipp, Keutzer, Candido, Uyttendaele, and Darrell]{reed2023scale}
Colorado~J Reed, Ritwik Gupta, Shufan Li, Sarah Brockman, Christopher Funk, Brian Clipp, Kurt Keutzer, Salvatore Candido, Matt Uyttendaele, and Trevor Darrell.
\newblock {Scale-MAE}: A scale-aware masked autoencoder for multiscale geospatial representation learning.
\newblock In \emph{Proceedings of the IEEE/CVF International Conference on Computer Vision}, pages 4088--4099, 2023.

\bibitem[Reichstein et~al.(2019)Reichstein, Camps-Valls, Stevens, Jung, Denzler, Carvalhais, and Prabhat]{reichstein2019deep}
Markus Reichstein, Gustau Camps-Valls, Bjorn Stevens, Martin Jung, Joachim Denzler, Nuno Carvalhais, and Prabhat.
\newblock Deep learning and process understanding for data-driven {Earth} system science.
\newblock \emph{Nature}, 566\penalty0 (7743):\penalty0 195--204, 2019.

\bibitem[Steiner et~al.(2021)Steiner, Kolesnikov, Zhai, Wightman, Uszkoreit, and Beyer]{steiner2021train}
Andreas Steiner, Alexander Kolesnikov, Xiaohua Zhai, Ross Wightman, Jakob Uszkoreit, and Lucas Beyer.
\newblock How to train your {ViT}? data, augmentation, and regularization in vision transformers.
\newblock \emph{arXiv preprint arXiv:2106.10270}, 2021.

\bibitem[Sumbul et~al.(2025)Sumbul, Xu, Dalsasso, and Tuia]{sumbul2025smarties}
Gencer Sumbul, Chang Xu, Emanuele Dalsasso, and Devis Tuia.
\newblock Smarties: Spectrum-aware multi-sensor auto-encoder for remote sensing images.
\newblock \emph{arXiv preprint arXiv:2506.19585}, 2025.

\bibitem[Tang et~al.(2024)Tang, Cozma, Georgiou, and Qi]{tang2024cross}
Maofeng Tang, Andrei Cozma, Konstantinos Georgiou, and Hairong Qi.
\newblock {Cross-Scale MAE}: A tale of multiscale exploitation in remote sensing.
\newblock \emph{Advances in Neural Information Processing Systems}, 36, 2024.

\bibitem[Tao et~al.(2023)Tao, Qi, Zhang, Zhu, Lu, and Li]{tao2023tov}
Chao Tao, Ji~Qi, Guo Zhang, Qing Zhu, Weipeng Lu, and Haifeng Li.
\newblock Tov: The original vision model for optical remote sensing image understanding via self-supervised learning.
\newblock \emph{IEEE Journal of Selected Topics in Applied Earth Observations and Remote Sensing}, 16:\penalty0 4916--4930, 2023.

\bibitem[Tseng et~al.(2025)Tseng, Fuller, Reil, Herzog, Beukema, Bastani, Green, Shelhamer, Kerner, and Rolnick]{tseng2025galileo}
Gabriel Tseng, Anthony Fuller, Marlena Reil, Henry Herzog, Patrick Beukema, Favyen Bastani, James~R. Green, Evan Shelhamer, Hannah Kerner, and David Rolnick.
\newblock Galileo: Learning global \& local features of many remote sensing modalities.
\newblock \emph{arXiv preprint arXiv:2502.09356}, 2025.

\bibitem[Van~der Maaten and Hinton(2008)]{van2008visualizing}
Laurens Van~der Maaten and Geoffrey Hinton.
\newblock Visualizing data using {t-SNE}.
\newblock \emph{Journal of Machine Learning Research}, 9\penalty0 (11), 2008.

\bibitem[Vaswani et~al.(2017)Vaswani, Shazeer, Parmar, Uszkoreit, Jones, Gomez, Kaiser, and Polosukhin]{vaswani2017attention}
Ashish Vaswani, Noam Shazeer, Niki Parmar, Jakob Uszkoreit, Llion Jones, Aidan~N Gomez, {\L}ukasz Kaiser, and Illia Polosukhin.
\newblock Attention is all you need.
\newblock \emph{Advances in Neural Information Processing Systems}, 30, 2017.

\bibitem[Waldmann et~al.(2025)Waldmann, Shah, Wang, Lehmann, Stewart, Xiong, Zhu, Bauer, and Chuang]{waldmann2025panopticon}
Leonard Waldmann, Ando Shah, Yi~Wang, Nils Lehmann, Adam~J. Stewart, Zhitong Xiong, Xiao~Xiang Zhu, Stefan Bauer, and John Chuang.
\newblock Panopticon: Advancing any-sensor foundation models for earth observation.
\newblock \emph{arXiv preprint arXiv:2503.10845}, 2025.

\bibitem[Wang et~al.(2022{\natexlab{a}})Wang, Zhang, Xu, Zhang, Du, Tao, and Zhang]{wang2022advancing}
Di~Wang, Qiming Zhang, Yufei Xu, Jing Zhang, Bo~Du, Dacheng Tao, and Liangpei Zhang.
\newblock Advancing plain vision transformer toward remote sensing foundation model.
\newblock \emph{IEEE Transactions on Geoscience and Remote Sensing}, 61:\penalty0 1--15, 2022{\natexlab{a}}.

\bibitem[Wang et~al.(2022{\natexlab{b}})Wang, Braham, Xiong, Liu, Albrecht, and Zhu]{wang2022ssl4eo}
Yi~Wang, Nassim Ait~Ali Braham, Zhitong Xiong, Chenying Liu, Conrad~M Albrecht, and Xiao~Xiang Zhu.
\newblock {SSL4EO-S12}: A large-scale multi-modal, multi-temporal dataset for self-supervised learning in {Earth} observation.
\newblock \emph{arXiv preprint arXiv:2211.07044}, 2022{\natexlab{b}}.

\bibitem[Wang et~al.(2023{\natexlab{a}})Wang, Albrecht, Braham, Liu, Xiong, and Zhu]{wang2023decur}
Yi~Wang, Conrad~M Albrecht, Nassim Ait~Ali Braham, Chenying Liu, Zhitong Xiong, and Xiao~Xiang Zhu.
\newblock {DeCUR}: decoupling common \& unique representations for multimodal self-supervision.
\newblock \emph{arXiv preprint arXiv:2309.05300}, 2023{\natexlab{a}}.

\bibitem[Wang et~al.(2023{\natexlab{b}})Wang, Braham, Xiong, Liu, Albrecht, and Zhu]{wang2023ssl4eo}
Yi~Wang, Nassim Ait~Ali Braham, Zhitong Xiong, Chenying Liu, Conrad~M Albrecht, and Xiao~Xiang Zhu.
\newblock {SSL4EO-S12}: A large-scale multimodal, multitemporal dataset for self-supervised learning in {Earth} observation.
\newblock \emph{IEEE Geoscience and Remote Sensing Magazine}, 11\penalty0 (3):\penalty0 98--106, 2023{\natexlab{b}}.

\bibitem[Wang et~al.(2023{\natexlab{c}})Wang, Hern{\'a}ndez, Albrecht, and Zhu]{wang2023feature}
Yi~Wang, Hugo~Hern{\'a}ndez Hern{\'a}ndez, Conrad~M Albrecht, and Xiao~Xiang Zhu.
\newblock Feature guided masked autoencoder for self-supervised learning in remote sensing.
\newblock \emph{arXiv preprint arXiv:2310.18653}, 2023{\natexlab{c}}.

\bibitem[Xiao et~al.(2018)Xiao, Liu, Zhou, Jiang, and Sun]{xiao2018unified}
Tete Xiao, Yingcheng Liu, Bolei Zhou, Yuning Jiang, and Jian Sun.
\newblock Unified perceptual parsing for scene understanding.
\newblock In \emph{Proceedings of the European Conference on Computer Vision (ECCV)}, pages 418--434, 2018.

\bibitem[Xiong et~al.(2024)Xiong, Wang, Zhang, and Zhu]{OFA}
Zhitong Xiong, Yi~Wang, Fahong Zhang, and Xiao~Xiang Zhu.
\newblock One for all: Toward unified foundation models for {Earth} vision.
\newblock \emph{arXiv preprint arXiv:2401.07527}, 2024.

\bibitem[Yao et~al.(2023)Yao, Lu, Yang, Xu, Liu, Hu, Yu, Liu, Deng, Tang, et~al.]{yao2023ringmo}
Fanglong Yao, Wanxuan Lu, Heming Yang, Liangyu Xu, Chenglong Liu, Leiyi Hu, Hongfeng Yu, Nayu Liu, Chubo Deng, Deke Tang, et~al.
\newblock {RingMo-sense}: Remote sensing foundation model for spatiotemporal prediction via spatiotemporal evolution disentangling.
\newblock \emph{IEEE Transactions on Geoscience and Remote Sensing}, 2023.

\bibitem[Zhang et~al.(2025)Zhang, Ru, Wu, Yu, Liang, Li, and Chen]{zhang2025skysensev2}
Yingying Zhang, Lixiang Ru, Kang Wu, Lei Yu, Lei Liang, Yansheng Li, and Jingdong Chen.
\newblock Skysense v2: A unified foundation model for multi-modal remote sensing.
\newblock \emph{arXiv preprint arXiv:2507.13812}, 2025.

\bibitem[Zucker and Regehr(2002)]{zucker2002short}
Robert~S Zucker and Wade~G Regehr.
\newblock Short-term synaptic plasticity.
\newblock \emph{Annual Review of Physiology}, 64\penalty0 (1):\penalty0 355--405, 2002.

\end{thebibliography}

\end{document}